\pdfoutput=1

\documentclass[11pt]{article}

\usepackage{acl}

\usepackage{times}
\usepackage{latexsym}

\usepackage[T1]{fontenc}
\usepackage[utf8]{inputenc}

\usepackage{microtype}

\usepackage{graphics}
\usepackage{graphicx}
\usepackage{xcolor}
\usepackage{amsmath}
\usepackage{cleveref}
\usepackage{bbding}
\usepackage{enumitem}
\usepackage{wasysym}
\usepackage{amssymb}
\usepackage{booktabs}
\usepackage{subcaption}
\usepackage{array,multirow}
\usepackage{adjustbox}
\usepackage{longtable}
\usepackage{tikz-cd}
\usepackage{placeins}
\usepackage{float}
\usepackage{tablefootnote}
\usepackage{tocloft}
\newcommand{\tabitem}{~~\llap{\textbullet}~~}
\usepackage{booktabs, makecell}

\Crefname{section}{§}{§§}
\Crefname{section}{§}{§§}

\everypar{\looseness=-1}
  {\list{}{\leftmargin=0in\rightmargin=0in}\item[]}%
  {\endlist}

\title{It Takes Two to Tango: Navigating Conceptualizations of NLP Tasks and Measurements of Performance}

\author{Arjun Subramonian \\
  University of California, Los Angeles \\
  \texttt{arjunsub@cs.ucla.edu} \\
  \And
  Xingdi Yuan \\
  Microsoft Research \\
  \texttt{Eric.Yuan@microsoft.com} \\
  \AND
  Hal Daum\'e III \\
  University of Maryland \\
  Microsoft Research \\
  \texttt{me@hal3.name} \\
  \And
  Su Lin Blodgett \\
  Microsoft Research \\
  \texttt{SuLin.Blodgett@microsoft.com} \\
  {} \\
  }
\begin{document}

\maketitle
\begin{abstract}
Progress in NLP is increasingly measured through benchmarks; hence, contextualizing progress requires understanding when and why practitioners may disagree about the validity of benchmarks. We develop a taxonomy of disagreement, drawing on tools from measurement modeling, and distinguish between two types of disagreement: 1) how tasks are conceptualized and 2) how measurements of model performance are operationalized. To provide evidence for our taxonomy, we conduct a meta-analysis of relevant literature to understand how NLP tasks are conceptualized, as well as a survey of practitioners about their impressions of different factors that affect benchmark validity. Our meta-analysis and survey across eight tasks, ranging from coreference resolution to question answering, uncover that tasks are generally not clearly and consistently conceptualized and benchmarks suffer from operationalization disagreements. These findings support our proposed taxonomy of disagreement. Finally, based on our taxonomy, we present a framework for constructing benchmarks and documenting their limitations.
\end{abstract}

\section{Introduction}

Claims of progress in NLP are often premised on how models perform on benchmarks for various NLP tasks\footnote{We disambiguate ``benchmarks'' and ``tasks'' in \Cref{sec:task_disambiguation}.} (e.g., coreference resolution, question answering) \cite{Wang2018GLUEAM, Wang2019SuperGLUEAS, Hu2020XTREMEAM, gehrmann-etal-2021-gem}. Benchmarks instantiate a task with a specific format, dataset of correct input-output pairs, and an evaluation metric \cite{Bowman2021WhatWI}, and they are intended to serve as measurement models for  performance on the task. 
On the one hand, benchmarks allow for performance results to be easily compared across a rapidly-rising number of NLP models \cite{schlangen-2021-targeting, ruder2021}.
Additionally, many NLP benchmarks are easily accessible via open-source platforms \cite{Lhoest2021DatasetsAC}, which reduces the need of  practitioners to construct new evaluation datasets and metrics from scratch.
However, prior research has identified numerous threats to the validity of benchmarks (i.e., how well benchmarks assess the ability of models to correctly perform tasks). These threats include spurious correlations and poorly-aligned metrics (refer to \Cref{tab:benchmark_issues} in the appendix).

However, little literature has surfaced sources of \textit{disagreement} among NLP practitioners about benchmark validity, which is paramount to contextualize progress in the field. Hence, we develop a taxonomy of disagreement based on measurement modeling (from the social sciences \cite{Adcock2001MeasurementVA, Jacobs2021MeasurementAF}). Our taxonomy critically distinguishes between disagreement in how tasks are conceptualized and how measurements of model performance are operationalized \cite{Blodgett2021StereotypingNS}. It thereby goes beyond prior examinations of NLP benchmarking methodology, which assume that tasks are generally clearly and consistently understood from person to person \cite{schlangen-2021-targeting, Bowman2021WhatWI}. This is important because our taxonomy captures that practitioners may perceive a benchmark for a task to have poor validity because they conceptualize the task differently than the benchmark creators do, and not simply because of the creators' oversights or mechanistic failures when constructing the benchmark. (We validate this hypothesis empirically in \Cref{sec:results_task_conceptualization}.) Furthermore, our taxonomy addresses that benchmarks can shape practitioners' conceptualization of a task.

\begin{figure*}[!ht]
    \adjustbox{scale=0.6,center}{%
    \begin{tikzcd}
    	&& {\text{Benchmark Disagreements}} \\
    	\\
    	& \parbox{3cm}{\centering Conceptualization Disagreements %
     }
            && \parbox{3cm}{\centering Face Validity}
            & \parbox{3cm}{\centering Operationalization Disagreements}
            & \parbox{3cm}{\centering Consequential Validity} \\
    	\\
    	\parbox{3cm}{\centering Model Capabilities $(C_\tau)$}
     & \parbox{3cm}{\centering Performance Correctness $ (y_\tau, M_\tau, \neg M_\tau)$}
     & \parbox{3cm}{\centering Essentially Contested Constructs $(E_\tau)$}
     & \parbox{3cm}{\centering Substantive Validity}
     & \parbox{3cm}{\centering Discriminant Validity}
     & \parbox{3cm}{\centering Convergent Validity}
    	\arrow[from=1-3, to=3-2]
    	\arrow[from=3-2, to=5-1]
    	\arrow[from=3-2, to=5-2]
    	\arrow[from=3-2, to=5-3]
    	\arrow[from=3-5, to=3-4]
    	\arrow[from=3-5, to=5-4]
    	\arrow[from=3-5, to=5-5]
    	\arrow[from=3-5, to=5-6]
    	\arrow[from=3-5, to=3-6]
    	\arrow[from=3-5, to=3-4]
    	\arrow[from=1-3, to=3-5]
    \end{tikzcd}}
    \caption{Bird's eye view of our taxonomy comprising conceptualization and operationalization disagreements.}
    \label{fig:taxonomy_diagram}
\end{figure*}
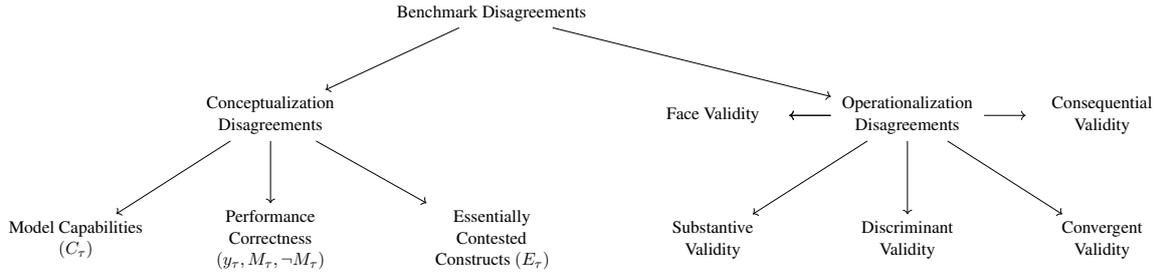

Ultimately, our taxonomy equips practitioners with a language to structure their thinking around and communicate their perceptions of benchmark validity. To provide evidence for our taxonomy, we conduct a survey of practitioners ($N$ = 46)
about their opinions on different factors that affect benchmark validity: how contested tasks are and the quality of common benchmark formats, datasets, and metrics for tasks. We further conduct a meta-analysis of relevant literature to understand how tasks are conceptualized. Our meta-analysis and survey across eight tasks, ranging from coreference resolution to question answering, uncover that tasks are generally not clearly and consistently conceptualized and benchmarks suffer from operationalization disagreements. These findings support our taxonomy of disagreement. Finally, based on our taxonomy, we present a framework for constructing benchmarks and documenting their limitations.

\section{Related Work}

\noindent \textbf{Community surveys} Researchers have conducted community surveys of NLP evaluation practices, often to surface perceptions that are not stated in related literature.
\citet{metasurvey2022} survey NLP practitioners to ``elicit opinions on controversial issues'' around benchmarking.
\citet{Zhou2022DeconstructingNE} survey NLG practitioners to uncover ``goals, community practices, assumptions, and constraints that shape NLG evaluations.''
\citet{dev-etal-2021-harms} survey non-binary individuals to understand how they are not included in NLP model bias evaluations.
We survey NLP practitioners to excavate perceptions of how contested tasks are and how well benchmarks measure model performance on tasks.

\noindent \textbf{Benchmark validity} A few previous works have studied benchmark validity through a measurement modeling lens \cite{Jacobs2021MeasurementAF}.
\citet{Blodgett2021StereotypingNS} analyze NLP bias evaluation benchmarks to inventory conceptualization and operationalization disagreements that threaten their validity as measurement models for stereotyping.
\citet{liao2021are} review papers from various machine learning subfields to characterize benchmarks from the angles of internal 
and external validity.
\citet{Raji2021AIAT} argue that benchmarks cannot measure ``progress towards general ability on vague tasks such as [...] `language understanding','' and hence lack construct validity.
We draw from measurement modeling to navigate how perceptions of validity issues with NLP benchmarks arise.

\raggedbottom

\section{Taxonomy of Disagreement}

We present our taxonomy of disagreement about the validity of NLP benchmarks (displayed in \Cref{fig:taxonomy_diagram}). Drawing from measurement modeling \cite{Jacobs2021MeasurementAF}, our taxonomy critically distinguishes between disagreement in: 1) how a task $\tau$ is conceptualized, and 2) how a benchmark $B_\tau$ operationalizes measurements of model performance on $\tau$.
We provide evidence for our taxonomy in \Cref{sec:results}, via our survey results and a meta-analysis of relevant literature.

\subsection{Task Conceptualization}
\label{sec:preliminaries_task_conceptualization}

$\tau$ is \textit{contested} when it lacks consistency or clarity in how it is conceptualized. In this case, because $B_\tau$ operationalizes measurements for model performance on $B_\tau$'s creators'\footnote{By ``creators,'' we refer to all individuals involved in the construction of $B_\tau$, including crowdworkers. We do not claim that all the creators of $B_\tau$ necessarily have nor does $B_\tau$ necessarily encode a consistent conceptualization of $\tau$. For example, the Universal Dependencies Treebank attempts to consolidate different conceptualizations of dependency parsing \cite{nivre-etal-2016-universal}; hence,
it likely fails to exactly match any individual linguist's conceptualization of syntax.}
conceptualization of $\tau$, there will necessarily be disagreement about the content validity of $B_\tau$ \cite{Jacobs2021MeasurementAF}. Disagreement in $\tau$'s conceptualization can stem from the following constructs with which $\tau$ is inextricably entangled:

\begin{itemize}[noitemsep,topsep=0pt,parsep=0pt,partopsep=0pt,leftmargin=*]

\item \textbf{Model capabilities:} Practitioners may disagree or lack clarity on the set of model capabilities $C_\tau$ that they assume $\tau$ involves \cite{Gardner2019QuestionAI, ribeiro-etal-2020-beyond, schlangen-2021-targeting}. Our conceptualization of $C_\tau$ is broader than ``cognitive capabilities'' \cite{Paullada2021DataAI}, encompassing e.g., handling various genres of text. However, $C_\tau$ can also include the coarse-grained capability of performing $\tau$ correctly. In contrast to \citet{schlangen-2021-targeting}, we argue that practitioners may determine $C_\tau$ in a top-down or bottom-up manner. They may first conceptualize $\tau$ as a specific real-world application and identify $C_\tau$ required to meet the needs of application users \cite{Cao2022WhatsDB}. Alternatively, practitioners may first identify $C_\tau$ that they believe to be linguistically interesting or crucial to general-purpose language systems, and subsequently devise $\tau$ such that $C_\tau$ is necessary to perform $\tau$ correctly \cite{Pericliev1984HandlingSA, schlangen-2021-targeting, Mahowald2023DissociatingLA}. In both cases, we gauge the extent to which a model possesses $C_\tau$ by proxy, by attempting to measure its performance on $\tau$.

\item \textbf{Performance correctness:} Practitioners may disagree or lack clarity on what constitutes performing $\tau$ correctly \cite{jamison-gurevych-2015-noise, Baan2022StopMC, Plank2022TheO}. This could include different perspectives on correct outputs $y_\tau$, as well as acceptable methods $M_\tau$ and unacceptable methods $\neg M_\tau$ for performing $\tau$ correctly \cite{Teney2022PredictingIN}.

\item \textbf{Essentially contested constructs:} Practitioners often disagree or lack clarity on essentially contested constructs $E_\tau$ entangled with $\tau$. A construct is essentially contested when its significance is generally understood, but there is frequent disagreement on what it looks like (e.g., language understanding, justice) \cite{gallie1955contested}. Developing criteria for whether a construct is essentially contested has been a subject of philosophical study for decades. For instance, \citet{gallie1955contested} posited that essenially contested constructs must have ``reciprocal recognition of their contested character among contending parties'' and ``an original exemplar that anchors conceptual meaning,'' among other characteristics \cite{Collier2006EssentiallyCC}. 

\end{itemize}

Model capabilities, performance correctness, and essentially contested constructs are mutually-building. $C_\tau$ (capabilities assumed to be involved to perform $\tau$ correctly) rely on a particular understanding of $y_\tau$. Similarly, $M_\tau$ (acceptable methods for performing $\tau$ correctly) may overlap with $C_\tau$.

\subsection{Perceptions of Benchmark Validity}
\label{sec:preliminaries_perceptions_of_benchmark_validity}

Our taxonomy connects disagreement in how $\tau$ is conceptualized to impressions of the validity of $B_\tau$ (i.e., how well $B_\tau$ operationalizes measurements of model performance on $\tau$). In particular, there are two reasons for perceptions of poor benchmark validity: disagreements in how the task is conceptualized, and operationalization disagreements. We delve into these reasons, with examples, in \Cref{sec:results}.

\begin{itemize}[noitemsep,topsep=0pt,parsep=0pt,partopsep=0pt,leftmargin=*]
    \item \textbf{Conceptualization disagreements:} Disagreements in how practitioners fundamentally conceptualize an aspect of $\tau$ (e.g., $C_\tau$, $y_\tau$, $M_\tau$, $\neg M_\tau$, $E_\tau$)  necessarily yields disagreements about the content validity of $B_\tau$. For example, \citet{williams-etal-2018-broad} construct MNLI because they conceptualize natural language inference as requiring models to handle various text genres, which they perceive SNLI ``falls short of providing a sufficient testing ground for'' because  ``sentences in SNLI are derived from only a single text genre.''
    Additionally, practitioners' conceptualizations of tasks can evolve over time, and even be influenced by the benchmarks with which they work. For instance, SQuAD arguably radically shifted practitioners' conceptualizations of QA from open-ended information retrieval to reading comprehension-style questions \cite{Rajpurkar2016SQuAD1Q}.
    As such, constructing valid benchmarks for a task can be a game with a shifting goalpost.
    
    \item \textbf{Operationalization disagreements:} Consider a set $P_{B_\tau}$ of practitioner(s) whose conceptualization of an aspect of $\tau$ aligns with that of the creators of $B_\tau$. Operationalization disagreements are choices made by the creators of $B_\tau$ (with respect to task format, dataset, and metric) that even within $P_{B_\tau}$, engender divergent perceptions of $B_\tau$'s validity. As an example, consider practitioners $P_{B_\tau}$ who believe that metrics for machine translation quality should ``yield judgments that correlate highly with human judgments'' \cite{NEURIPS2021_260c2432}. \citet{NEURIPS2021_260c2432}, motivated by their impression that popular automatic evaluation metrics in NLG (e.g., BLEU, ROUGE) ``weakly'' operationalize how humans judge machine translations, propose a new metric MAUVE.

\end{itemize}

We provide an extended discussion of conceptualization and operationalization disagreements in \Cref{sec:disagreements_supp}.

\section{Survey Methodology}
\label{sec:methods}

With our taxonomy in mind, we conduct a survey of NLP practitioners\footnote{Following \citet{Zhou2022DeconstructingNE}, by ``practitioners,'' we refer to academic and industry researchers, applied scientists, and engineers who have experience with NLP tasks or evaluating NLP models or systems.} ($N = 46$) to surface and understand for various NLP tasks, practitioners' perceptions of: \textbf{(1)} the extent to which the tasks appear to have a clear and consistent conceptualization and \textbf{(2)} the quality of benchmarks (with respect to task format, dataset, and metric). We ultimately chose to include the following tasks in our survey: Sentiment Analysis (\textsc{Sent}), Natural Language Inference (\textsc{NLI}), Question Answering (\textsc{QA}), Summarization (\textsc{Sum}), Machine Translation (\textsc{MT}), Named-Entity Recognition (\textsc{NER}), Coreference Resolution (\textsc{Coref}), and Dependency Parsing (\textsc{Dep}). We detail our task selection protocol in \Cref{sec:survey_task_selection}. 

\noindent \textbf{Survey topics} In our survey, we begin by asking participants about their background (i.e., occupation and experience with NLP) to understand the demographics of our sample. We then inquire into participants' initial impressions of how current state-of-the-art
NLP models perform on various NLP tasks; we do this prior to asking participants to engage more critically with task definitions and benchmarks, so as not to sway their responses. Subsequently, for each task, we ask participants about their familiarity with the task, and if they are familiar, their perceptions of the \textbf{(a)} clarity and consistency of the task's definition or conceptualization, \textbf{(b)} extent to which common task formats capture the underlying language-related skill, \textbf{(c)} quality of benchmark datasets and metrics, and \textbf{(d)} progress on the task. We utilize perceptions of \textbf{(a)} as a proxy for how contested tasks are across practitioners. We do this because it is not feasible to collect and compare participants' raw task conceptualizations in a quantifiable manner.
Furthermore, we collect perceptions of \textbf{(b)} and \textbf{(c)} to capture conceptualization and operationalization disagreements across benchmarks generally. We do not inquire into participants' impressions of \textbf{(b)} and \textbf{(c)} for specific benchmarks in order to keep the survey reasonably long and have a sufficient sample size.

For all survey questions that ask participants to rate their perception, we provide them with a scale that ranges from 1 to 6 with articulations of what 1 and 6 mean in the context of the question. We include the entirety of our survey and survey results in \Cref{sec:survey_questions}, and discuss participant guidance in \Cref{sec:participant_guidance}.

\begin{table}[t]
\small
\centering
\begin{tabular}{l|c}
\toprule
\textbf{Role} & \textbf{\#}\\
\midrule
Works on deployed systems & 6 \\
Industry practitioner (not researcher) & 7 \\
Industry researcher & 10 \\
Academic researcher & 32 \\
\bottomrule
\end{tabular}
\caption{Demographics of survey participants. Some participants identified with more than one role.}
\label{tab:survey_demographics}
\end{table}

\noindent \textbf{Survey recruitment and quality control} As seen in \Cref{tab:survey_demographics}, our sample is heavily skewed towards academic researchers; we detail our participant recruitment protocol and IRB approval in \Cref{sec:survey_participant_recruitment_irb}. We additionally document our quality control measures in \Cref{sec:survey_quality_control}.

\section{Results}
\label{sec:results}

\subsection{Task Conceptualization}
\label{sec:results_task_conceptualization}

\begin{figure}[!t]
    \centering
    \includegraphics[width=\columnwidth]{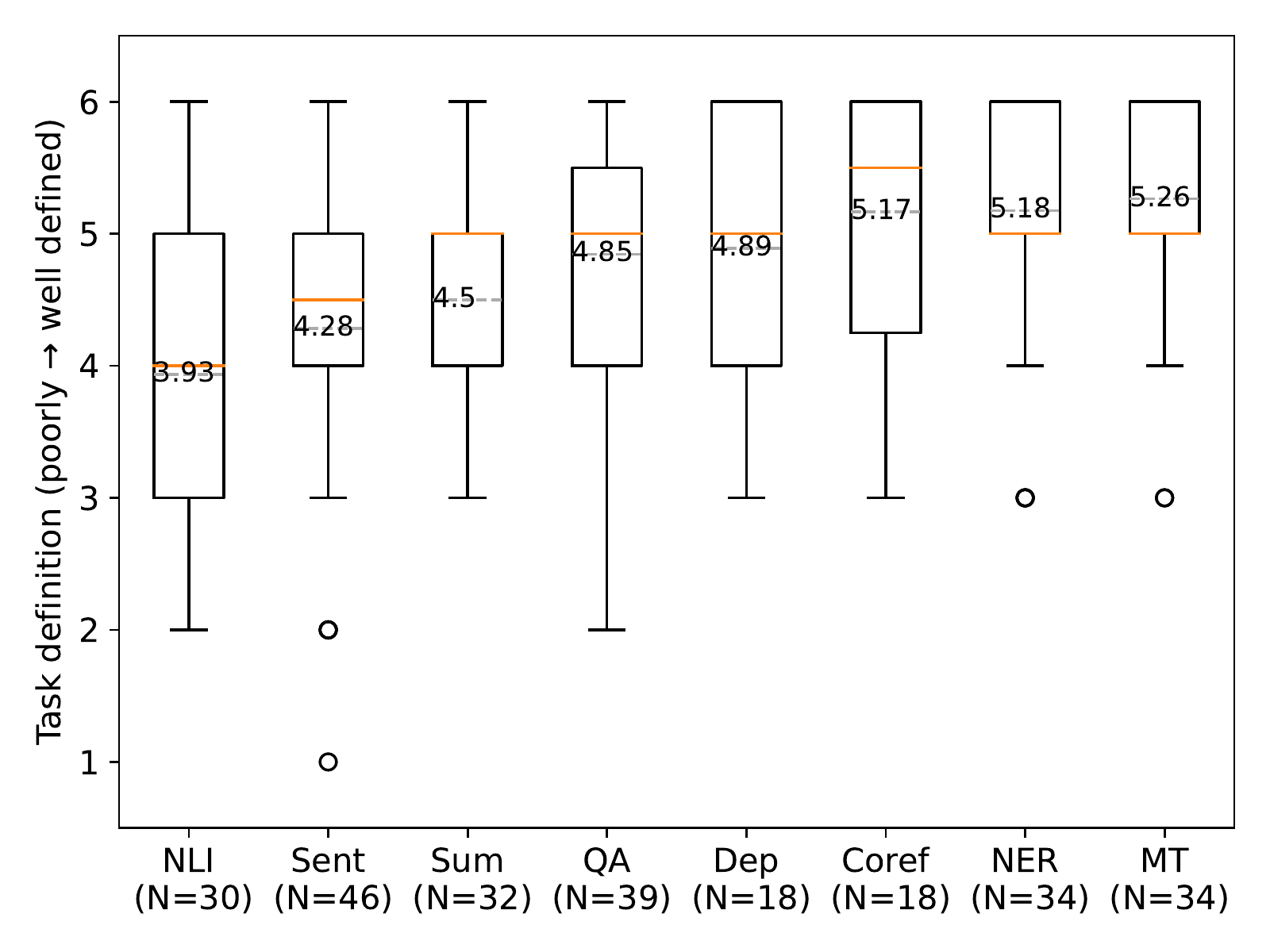}
    \caption{Perceived clarity and consistency of task definition. Orange lines indicate median score, while dashed lines indicate average score. %
    }
    \label{fig:task_definition_all_main}
\end{figure}

\Cref{fig:task_definition_all_main} shows how survey participants perceive the clarity and consistency with which various NLP tasks are conceptualized. We observe that:
\begin{itemize}[noitemsep,topsep=0pt,parsep=0pt,partopsep=0pt,leftmargin=*]
    \item \textbf{Tasks are not perfectly clearly or consistently conceptualized.} No task in \Cref{fig:task_definition_all_main} received a score of 6 from all participants.
    \item \textbf{Tasks are conceptualized with varying levels of clarity and consistency.} The tasks in \Cref{fig:task_definition_all_main} exhibit a range of average and median conceptualization scores. \textsc{NLI} and \textsc{Sent} appear to have objectives that are less clearly and consistently understood by practitioners, while \textsc{Coref} and \textsc{MT} seem to be better defined.
    \item \textbf{Practitioners diverge in their impressions of how clearly and consistently the NLP community conceptualizes a task.} Many tasks in \Cref{fig:task_definition_all_main} have a large interquartile range, and for \textsc{NLI} and \textsc{Sent}, scores span from 2 to 6.
\end{itemize}

To further provide evidence for these observations, we leverage our taxonomy (in particular, the sources of disagreement in task conceptualization described in  \Cref{sec:preliminaries_task_conceptualization}) and relevant literature.

\noindent \textbf{Model capabilities} In order to understand disagreement about involved capabilities $C_\tau$ for the tasks, we meta-analyze benchmarks that survey participants mention.  %
Specifically, for each task, we first select the 2--4
most frequently mentioned benchmarks; we then perform light open coding\footnote{Open coding refers to ``labeling concepts, defining and developing categories based on their properties and dimensions'' without a predefined list of categories \cite{KhandkarOC}.} on the papers that initially proposed these benchmarks in order to identify model capabilities\footnote{We restrict our attention to stated capabilities that lie below the surface of the capability of performing the task correctly \cite{schlangen-2021-targeting}. Some annotated datasets when proposed, were not intended for model evaluation, but were later repurposed as benchmarks (e.g., Penn Treebank \cite{Marcus1993BuildingAL}).} that the authors claim the benchmark assesses.

We find that for each task, stated capabilities overlap but often vary across benchmarks, suggesting disagreement in task conceptualization. For instance, for \textsc{Sum}, the authors of XSum claim that the benchmark assesses whether models can generate novel language, handle linguistic phenomena, and handle various domains \cite{Narayan2018DontGM}, while the authors of CNN/Daily Mail claim that this benchmark gauges whether models possess benchmark-external knowledge \cite{Nallapati2016AbstractiveTS}; however, authors of both benchmarks intend to test language understanding. We present all our meta-analysis results in \Cref{sec:model_capabilities}.

\noindent \textbf{Performance correctness} Correct outputs $y_\tau$ for a task may be inherently disagreed upon or unclear. For instance, in \textsc{MT}, the adequacy of translations in $y_\tau$ is subjective \cite{white-oconnell-1993-evaluation}; further, it can be unclear how to translate lexical and syntactic ambiguity in the source language \cite{Pericliev1984HandlingSA, Baker1994CopingWA}, or translate from a language without to with grammatical gender \cite{gonen-webster-2020-automatically}. We present additional examples in \Cref{tab:findings_summary}.

Practitioners can also disagree about acceptable methods $M_\tau$ and unacceptable methods $\neg M_\tau$ for performing a task correctly. For example, \citet{Sugawara_Stenetorp_Inui_Aizawa_2020} expect models to take certain actions when performing reading comprehension, e.g., \{recognize word order, resolve pronoun coreferences\} $\subset M_\tau$. On the other hand, numerous works have raised concerns about models exploiting annotation artifacts in \textsc{NLI} \cite{gururangan-etal-2018-annotation, poliak-etal-2018-hypothesis} and \textsc{QA} benchmarks \cite{Si2019WhatDB, Kavumba2019WhenCP, Chen2019UnderstandingDD}, which suggests that they view exploiting artifacts as part of $\neg M_\tau$.

\noindent \textbf{Essentially contested constructs} $E_\tau$ pose an issue when practitioners incorrectly presuppose that $E_\tau$ have clear and consistent conceptualizations, thus failing to communicate how they personally understand $E_\tau$. We present examples of essentially contested constructs $E_\tau$ entangled with various tasks in \Cref{tab:findings_summary}, elaborating on a few in this section. \textsc{Sent} presupposes that the essentially contested construct ``sentiment:'' 1) has a clear and consistent definition (e.g., falls on a spectrum between ``positive'' and ``negative''); 2) can be gleaned from text alone; and 3) admits expressions that are universally or predominantly interpreted the same way from person to person. However, there exists ``divergences of sentiments about different concepts'' across cultures \cite{Heise2014CulturalVI}, and hence ``sentiment'' $\in E_\tau$. Furthermore, \textsc{Coref}, in asking if two expressions refer to the same entity, presupposes that the essentially contested construct ``identity'' is clearly and consistently understood, and thus ``identity is never adequately defined'' \cite{recasens-etal-2010-typology}.

\begin{figure*}[!ht]
    \begin{subfigure}[]{0.33\textwidth}
        \centering
        \includegraphics[width=\textwidth]{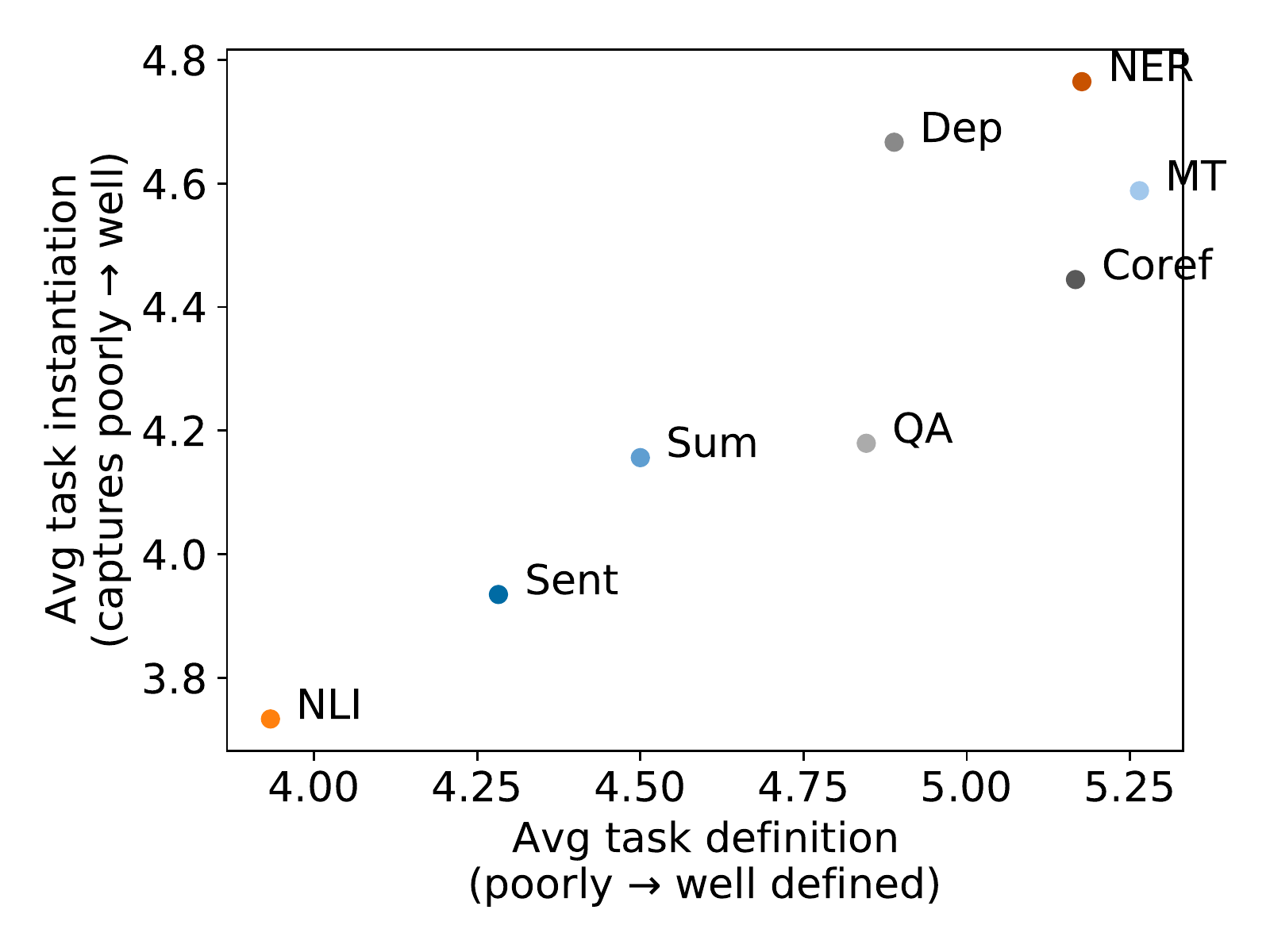}
        \caption{}
    \end{subfigure}
    \begin{subfigure}[]{0.33\textwidth}
        \centering
        \includegraphics[width=\textwidth]{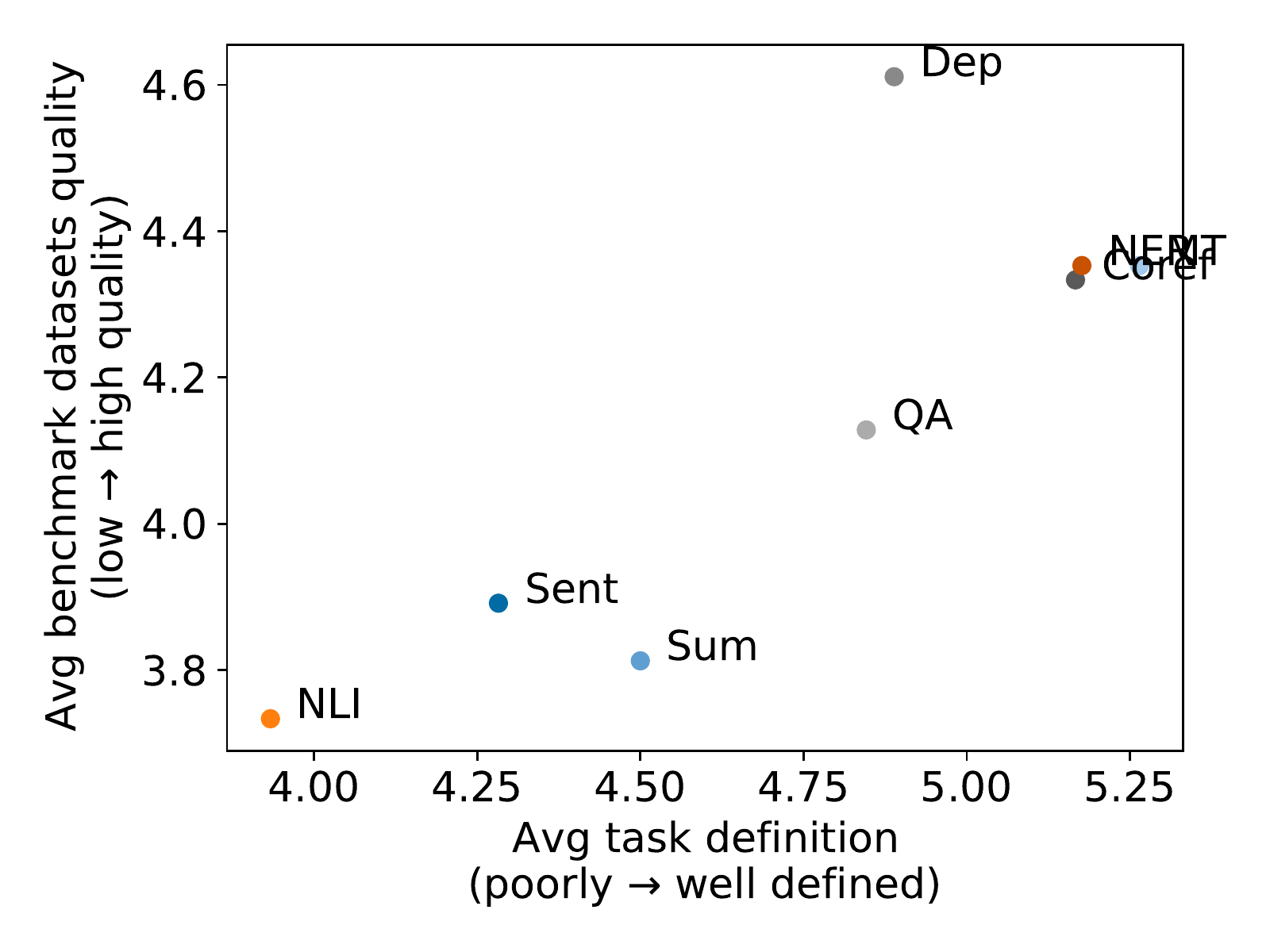}
        \caption{}
    \end{subfigure}
    \begin{subfigure}[]{0.33\textwidth}
        \centering
        \includegraphics[width=\textwidth]{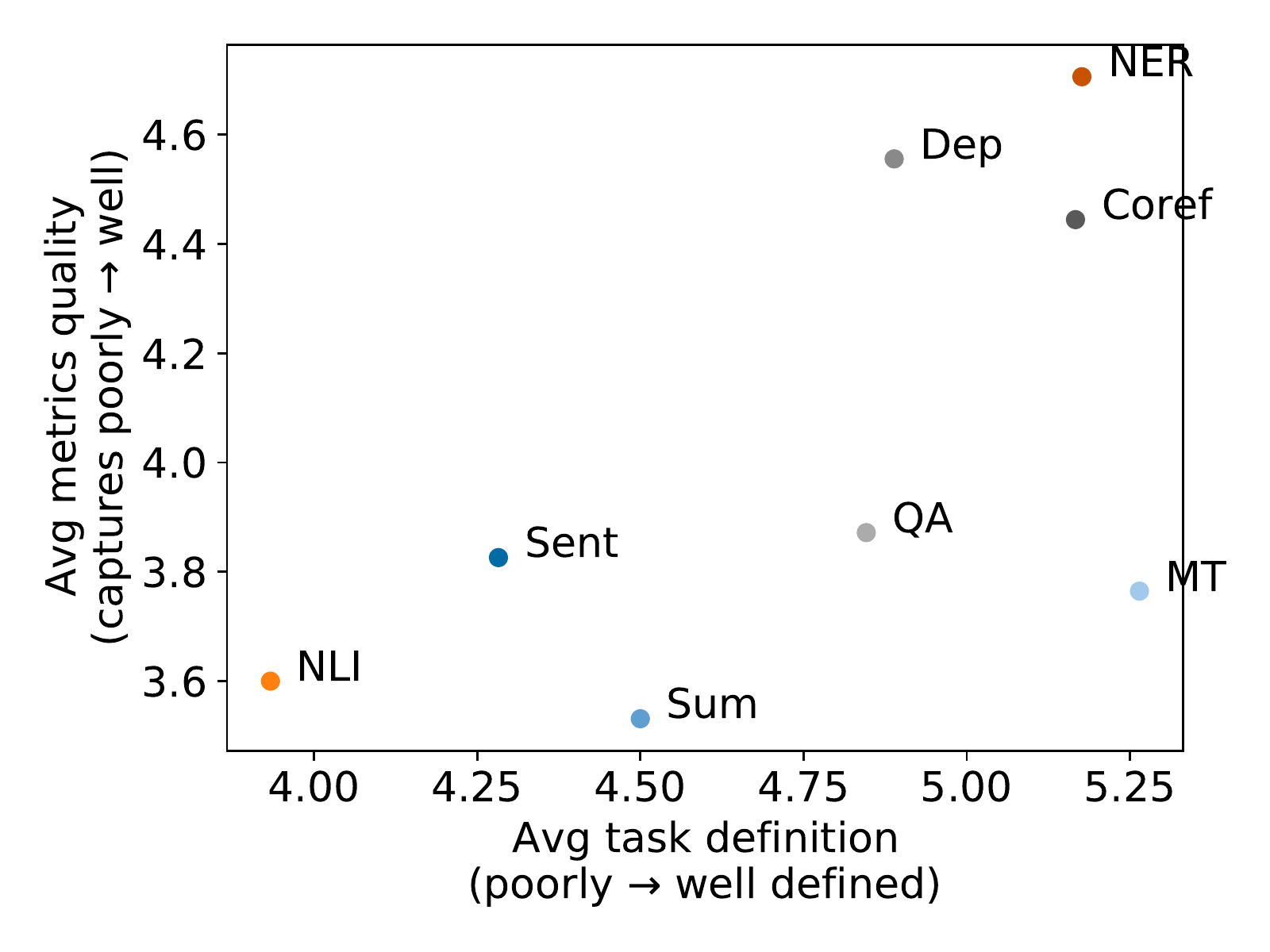}
        \caption{}
        \label{fig:benchmark_quality_metrics}
    \end{subfigure}
    \caption{Perceived quality of common task instantiations, benchmark datasets, and benchmark metrics vs. perceived clarity and consistency of task definition.}
    \label{fig:benchmark_quality_all}
\end{figure*}

Often, $C_\tau$ and $E_\tau$ overlap. As revealed by our meta-analysis of model capabilities, practitioners may believe that performing certain tasks involves:
\begin{itemize}[noitemsep,topsep=0pt,parsep=0pt,partopsep=0pt,leftmargin=*]
    \item \textbf{Possessing benchmark-external knowledge:} But, what constitutes benchmark-external knowledge is left ambiguous. For example, in \textsc{QA}, questions may involve ``commonsense knowledge'' \cite{Talmor2019CommonsenseQAAQ, schlegel-etal-2020-framework}, whose constitution is essentially unclear and inconsistently understood \cite{MUELLER2015339}. It is also unclear how much external knowledge and context \textsc{NER} requires to disambiguate entities \cite{Ratinov2009DesignCA}.
    \item \textbf{Being on par with humans:} However, practitioners often do not specify which humans (e.g., crowdworkers, trained syntacticians) with which they would like models to be on par, or use vague or problematic language in their specifications (e.g., ``normally-abled adults whose first language is English'' \cite{Levesque2011TheWS}). 
\end{itemize}

We discuss additional examples of essentially contested constructs in \Cref{sec:essentially_contested_examples}.

\subsection{Perceptions of Benchmark Validity}
\label{sec:examples_perceptions_of_benchmark_validity}
\Cref{fig:benchmark_quality_all} depicts for various tasks, how survey participants' perceptions of the quality of common benchmark task instantiations, datasets, \& metrics (which are central to benchmark validity) vary in relation to their perceptions of the clarity and consistency of how the task is defined. These plots show that there is generally a positive association between perceptions of benchmark validity and task contestedness. This observation indicates that benchmarks suffer from conceptualization disagreements. However, this observation could also reflect that NLP practitioners collapse task contestedness onto their perceptions of benchmark validity.

The plots also demonstrate that the association (especially between perceptions of metric quality and task contestedness) is weak, with seemingly well-defined tasks like \textsc{MT} facing impressions of low-quality metrics. This association weakness suggests that benchmarks suffer from operationalization disagreements. To provide evidence for our findings, we leverage relevant literature.

\noindent \textbf{Conceptualization disagreements} We describe some disagreements in the conceptualization of NLP tasks and provide examples of resultant conceptualization disagreements in \Cref{tab:findings_summary}.

\newcolumntype{P}[1]{>{\centering\arraybackslash}p{#1}}

\begin{table*}[!ht]
    \centering
    \begin{adjustbox}{max width=2.0\columnwidth}
    \begin{tabular}{|p{0.07\textwidth}|p{0.8\textwidth}|p{0.53\textwidth}|}
         \toprule
         \thead{\textbf{Task}} & \thead{\textbf{Disagreement in conceptualization?}} & \thead{\textbf{Disagreement examples}} \\
         \midrule
         \multirow{7}{=}{\textsc{NLI}} & $C_\tau$: yes (\Cref{tab:capabilities_measured_nli}).
                        & \multirow{6}{=}{SNLI, MNLI datasets operationalize validity of natural language inferences with single gold label \cite{Bowman2015ALA, williams-etal-2018-broad}.} \\
                        \cline{2-2}
                      & $y_\tau$: yes; inherent disagreement in validity of natural language inferences \cite{pavlick-kwiatkowski-2019-inherent}; lack of clarity and disagreement about $y_\tau$ when premise or hypothesis is question (\Cref{fig:mnli_example}). & \\
                       \cline{2-2}
                       & $E_\tau$: yes; \{understand language, possess benchmark-external knowledge\} $\subset E_\tau$ (\Cref{tab:capabilities_measured_nli}). & \\
         \midrule
         \multirow{6}{=}{\textsc{QA}} & $C_\tau$: yes (\Cref{tab:capabilities_measured_qa}).
                        & \multirow{4}{=}{HotpotQA, ReCoRD, MultiRC datasets operationalize reference answers with arbitrary precision \cite{schlegel-etal-2020-framework}.}
                        \\
                        \cline{2-2}
                     & $y_\tau$: yes; appropriate adequacy of answers in $y_\tau$ is subjective \cite{schlegel-etal-2020-framework}. & \\
                     \cline{2-2}
                     & $E_\tau$: yes; \{understand language, reason over a context, possess benchmark-external knowledge, be on par with humans\} $\subset E_\tau$  (\Cref{tab:capabilities_measured_qa}). & \\
        \midrule
        \multirow{7}{=}{\textsc{Coref}} & $C_\tau$: yes (\Cref{tab:capabilities_measured_coref}).
                                    & \multirow{5}{=}{OntoNotes dataset does not capture near-identity coreferences \cite{recasens-etal-2010-typology, Zeldes2022CanWF}.}
                                    \\
                     \cline{2-2}
                        & $y_\tau$: yes; inherent anaphoric ambiguity induces lack of clarity and disagreement about $y_\tau$ \cite{poesio-artstein-2005-reliability}. & \\
                         \cline{2-2}
                         & $E_\tau$: yes; \{identity, be on par with humans, possess benchmark-external knowledge\} $\subset E_\tau$ (\Cref{tab:capabilities_measured_coref}). & \\
        \midrule
        \multirow{7}{=}{\textsc{Sum}} & $C_\tau$: yes (\Cref{tab:capabilities_measured_sum}) .
                        & \multirow{5}{=}{benchmark datasets contain single gold summaries with varying levels of adequacy \cite{kano-etal-2021-quantifying}.}
                        \\
                        \cline{2-2}
                        & $y_\tau$: yes; ``goodness'' and adequacy of summaries in $y_\tau$ are subjective \cite{Nallapati2016AbstractiveTS, Li2021SubjectiveBI, ter-hoeve-etal-2022-makes}. & \\
                        \cline{2-2} 
                        & $E_\tau$: yes; \{ understand language, possess benchmark-external knowledge \} $\subset E_\tau$ (\Cref{tab:capabilities_measured_sum}). & \\
        \bottomrule
    \end{tabular}
    \end{adjustbox}
    \caption{Disagreements in the conceptualization of NLP tasks and relevant examples.}
    \label{tab:findings_summary}
\end{table*}

\noindent \textbf{Operationalization disagreements} Operationalization disagreements can be attributed to various factors. Measurement modeling naturally provides us with a language to categorize and discuss these factors, and in the process, theorize about the real world. Hence, we taxonomize operationalization disagreements through the lens of different threats to validity in the measurement modeling literature.

\begin{itemize}[noitemsep,topsep=0pt,parsep=0pt,partopsep=0pt,leftmargin=*]
\item \textbf{Face validity:} Benchmarks can have surface characteristics (e.g., incorrect or incomplete annotations) that affect perceptions of their quality. For instance, \textsc{QA}, \textsc{Coref}, and \textsc{NER} benchmarks often contain incorrect or incomplete annotations \cite{jie-etal-2019-better, schlegel-etal-2020-framework, Blodgett2021StereotypingNS}. Many \textsc{Sum} benchmarks have unfaithful reference summaries \cite{Zhang2022ExtractiveIN, Tang2022UnderstandingFE, Goyal2021AnnotatingAM}. \textsc{MT} benchmarks often contain incorrect reference translations \cite{Castilho2017ACQ}. 

\item \textbf{Substantive validity:} A benchmark may not exhaustively assess a model capability \cite{schlangen-2021-targeting}. For example, practitioners may conceptualize a task as involving the capability to handle phenomena in real-world data, but benchmark datasets (e.g., from ``constrained social media platforms'') can fail to ``reflect broader real-world phenomena'' \cite{Olteanu2016SocialDB, Hupkes2022StateoftheartGR}. For example, despite having saturated SST-2 \cite{Wang2019SuperGLUEAS}, NLP models struggle with domain shift, bi-polar words, negation \cite{Hussein2018ASO, Hossain2022AnAO}. Furthermore, \textsc{QA} benchmarks are often restricted to a single format (e.g., multiple-choice reading comprehension, story-cloze queries \cite{schlegel-etal-2020-framework}), which does not substantively instantiate \textsc{QA}. Moreover, the format of \textsc{MT} benchmarks (e.g., of W\textsc{MT} shared tasks) often precludes sufficient intersentential context for substantively assessing translations \cite{Toral2020ReassessingCO}.

\item \textbf{Discriminant validity:} Benchmarks may inadvertently assess undesired model capabilities or ``unacceptable'' methods of performing a task (e.g., picking up on spurious cues) \cite{Jacobs2021MeasurementAF}. For instance, despite having saturated SuperGLUE \textsc{NLI} benchmarks \cite{Wang2019SuperGLUEAS}, NLP models fail on a controlled evaluation set where it is not possible to rely on syntactic heuristics \cite{McCoy2019RightFT}.

\item \textbf{Convergent validity:} Benchmarks may not ``match other accepted measurements'' of performance on a task. For example, practitioners may consistently conceptualize \textsc{Sum} and \textsc{MT} as involving ``being on par with humans''; however, automatic evaluation metrics like ROUGE and BLEU are poorly aligned with human judgments of summarization \cite{deutsch-roth-2021-understanding, Deutsch2022ReExaminingSC} and translation \cite{Reiter2018ASR, Toral2020ReassessingCO, marie-etal-2021-scientific, amrhein2022aces} quality, respectively. This is reflected in \Cref{fig:benchmark_quality_metrics}, which shows that \textsc{Sum} and \textsc{MT} noticeably deviate from the positive trend; in particular, although these tasks are more consistently and clearly conceptualized, practitioners perceive their metrics to be low-quality (i.e., \textsc{Sum} and \textsc{MT} benchmarks have poor convergent validity).

\item \textbf{Consequential validity:} Practitioners may be concerned that the use of a benchmark has societal harms. For example, \textsc{Sent} benchmarks can reinforce hegemonic conceptions of emotion and and be culturally discriminatory \cite{emotion-dangers-2022}.  

\end{itemize}

Benchmark issues may threaten more than a single aspect of validity.

\begin{figure}[!hb]
    \centering
    \includegraphics[width=\columnwidth]{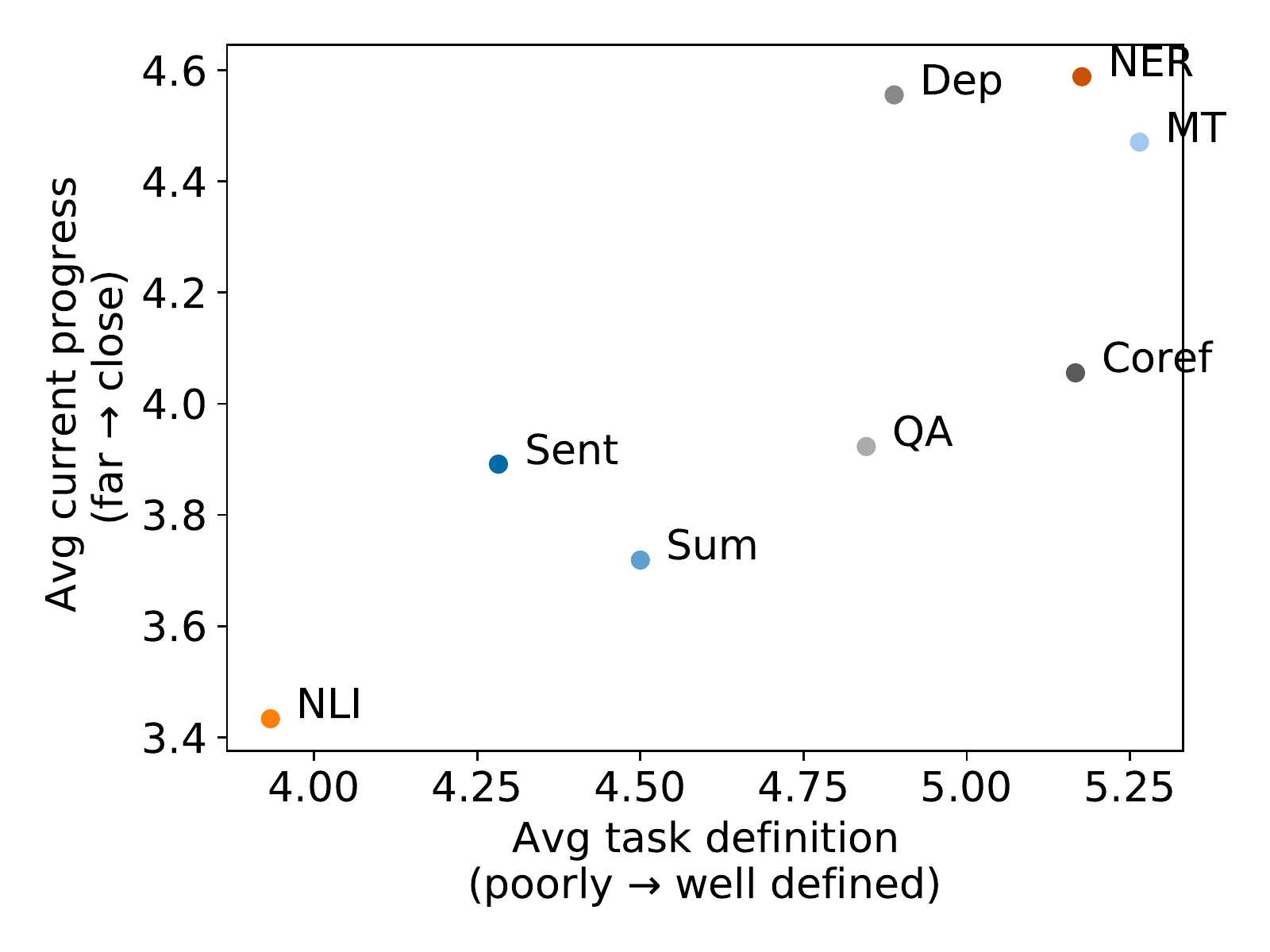}
    \caption{Perceived quality of common task instantiations, benchmark datasets, and benchmark metrics vs. perceived current progress on task.}
    \label{fig:current_progress_all_main}
\end{figure}

\raggedbottom

\begin{figure*}
    \begin{subfigure}[]{0.33\textwidth}
        \centering
        \includegraphics[width=\textwidth]{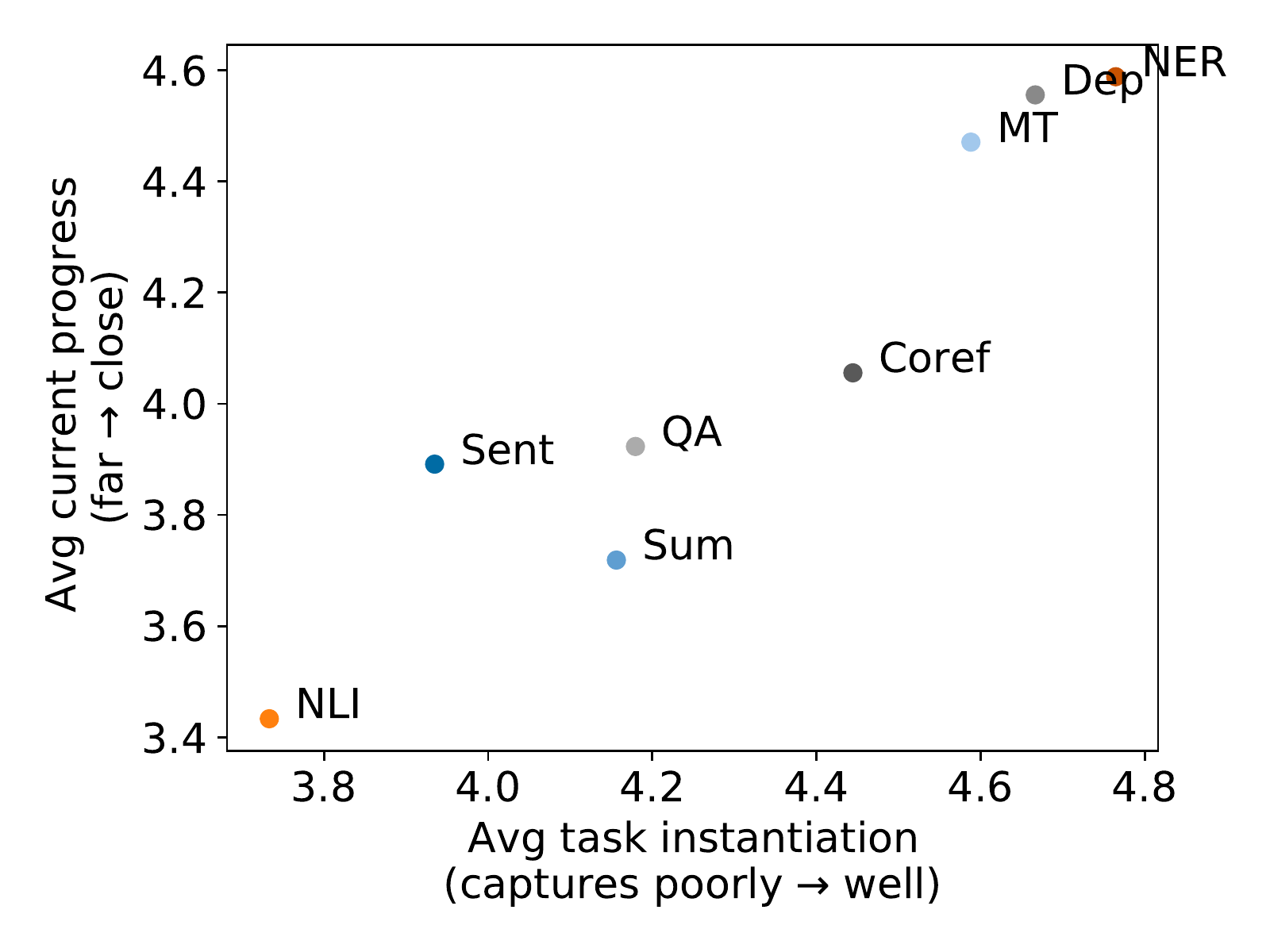}
        \caption{}
    \end{subfigure}
    \begin{subfigure}[]{0.33\textwidth}
        \centering
        \includegraphics[width=\textwidth]{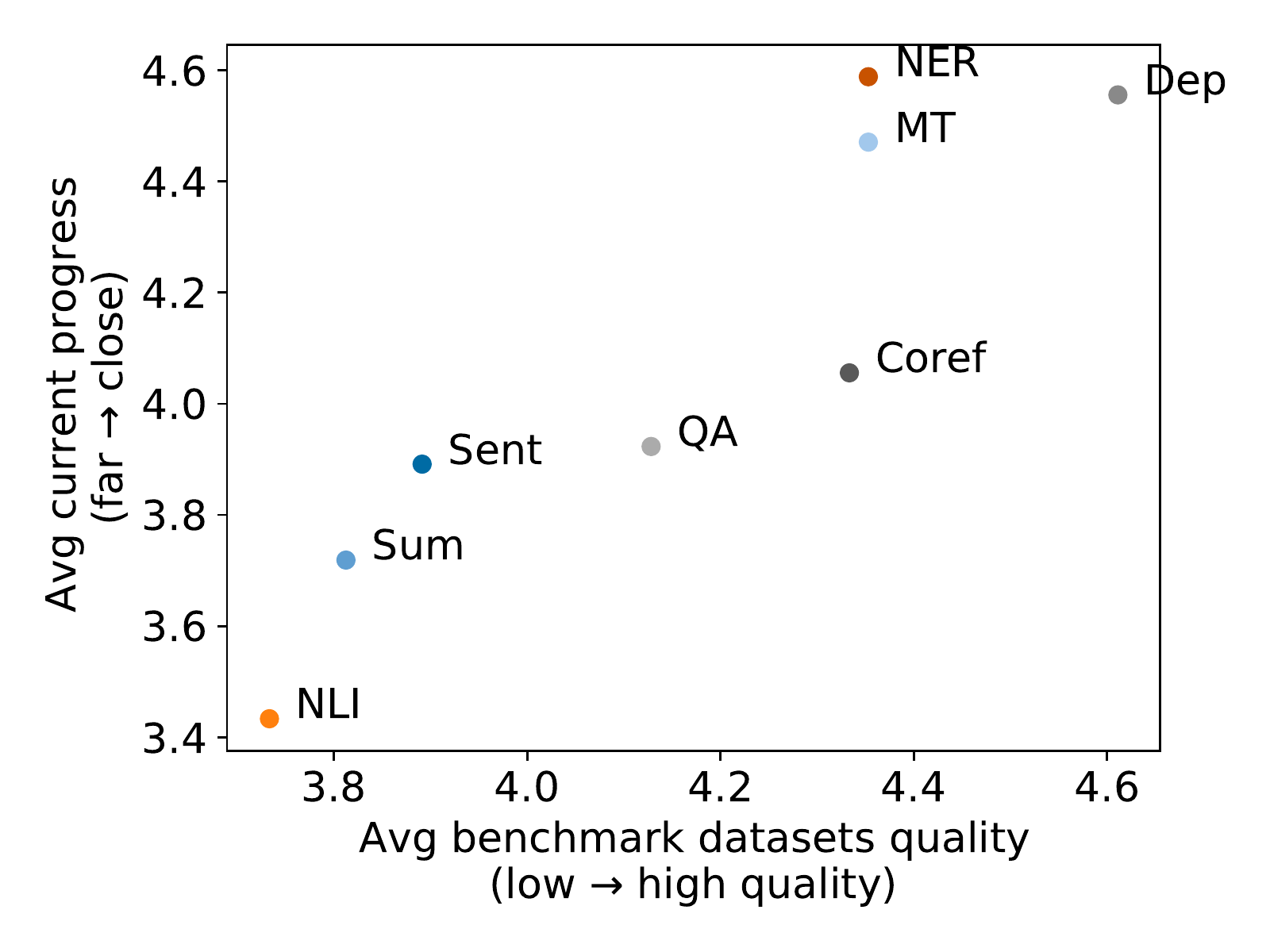}
        \caption{}
    \end{subfigure}
    \begin{subfigure}[]{0.33\textwidth}
        \centering
        \includegraphics[width=\textwidth]{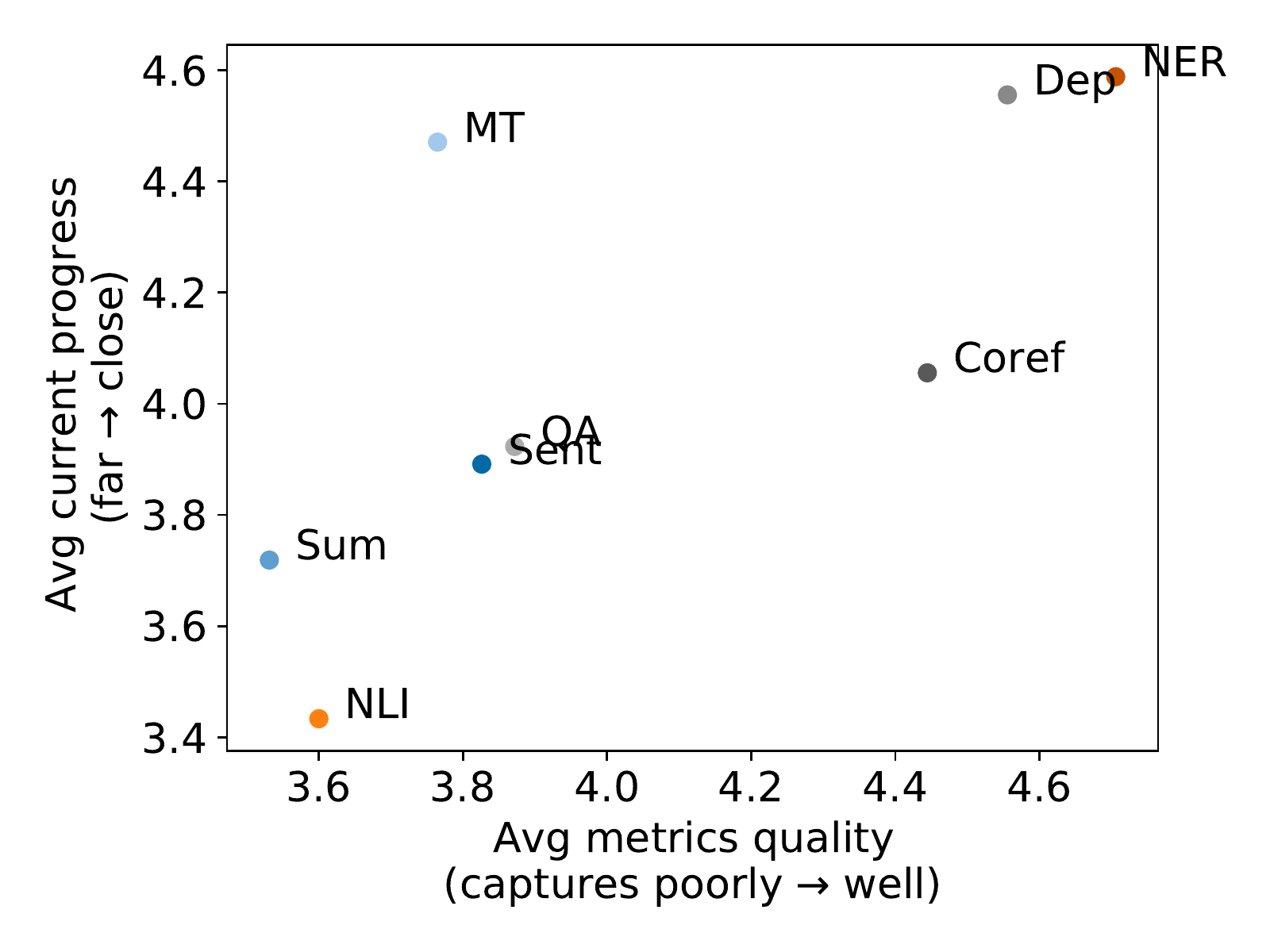}
        \caption{}
    \end{subfigure}
    \caption{Quality of common task instantiations, benchmark datasets, and benchmark metrics vs. perceived current progress on task among all responding practitioners.}
    \label{fig:current_progress_all_supp}
\end{figure*}

\subsection{Progress}

\Cref{fig:current_progress_all_main} (and \Cref{fig:current_progress_all_supp} in the appendix) suggest that perceptions of better task conceptualization and benchmark validity are associated with perceptions of stronger progress on the task. In reality, impressions of progress in NLP (especially for non-practitioners) may be disconnected from the validity of the benchmarks used to make claims about progress \cite{Bender2021OnTD}. This is important because claims of progress shape the social and academic capital of NLP, and are implicitly embedded in every research artifact, including scientific publications. Thus, towards proper science and accountability, NLP practitioners ought to make realistic and tenable claims about progress, and not overhype NLP models. Furthermore, progress is neither monolithic nor does it increase monotonically; it is critical to be transparent about benchmark validity issues and their implications for claims of progress. We must simultaneously re-imagine ``progress'' in NLP to encapsulate measuring, alleviating, and communicating benchmark validity issues.

\begin{table*}[!ht]
\small
\centering
\begin{tabular}{|p{0.95\textwidth}|}
\hline
\textbf{Conceptualization questions}\\
\textbf{Model capabilities:} Which $C_\tau$ do you believe $\tau$ involves and why? (e.g., Table 1 in \citet{ribeiro-etal-2020-beyond}) How does $C_\tau$ differ from the capabilities that other benchmarks for $\tau$ are intended to assess? \\
\textbf{Performance correctness:} How may $y_\tau, M_\tau, \neg M_\tau$ be contested? How did you involve relevant communities to co-create $B_\tau$? How would you accurately characterize ``solving'' $\tau$? \\
\textbf{Essentially contested constructs:} Do you define any $E_\tau$ (e.g., model capabilities) entangled with $\tau$? (e.g., ``universality'' in \citet{Bhatt2021OnTU}) How did you come up with the name of $\tau$ and $B_\tau$? Do you avoid employing overloaded or overclaiming terminology in your $\tau$'s name \cite{Shanahan2022TalkingAL}? \\
\textbf{Overarching questions:} How may $B_\tau$ limit ``progress'' to only working on one conceptualization of $\tau$? Do you hold space for others to propose alternatives? \\
\hline
\textbf{Operationalization questions} \\
\textbf{Validity:} How well does $B_\tau$ operationalize a measurement model for model performance on your conceptualization of $\tau$? What kinds of validity may $B_\tau$ lack and why? If $B_\tau$ were to indicate that a model performs exceptionally well on it, what can the NLP community conclude? \\
\hline
\end{tabular}
\caption{Documentation questions to facilitate the creation of NLP benchmarks.}
\label{tab:documentation_questions}
\end{table*}

\section{A Framework for NLP Benchmarks} 
\label{sec:new_benchmarks}
\label{sec:documentation}

Towards better documenting benchmarks' conceptualization and operationalization, we encourage benchmark creators to answer the questions in \Cref{tab:documentation_questions} in their future directions or limitations section when they propose a new benchmark $B_\tau$ for a task $\tau$. This framework is not a post-hoc intervention. We intend for benchmark creators to answer these questions before, during, and after they construct benchmarks; this framework should be grounded in care for and facilitating collective progress in NLP. Furthermore, creators should share their answers to these questions, so that this framework becomes normalized and shapes people’s thinking about their own contributions. Moreover, this framework is intended to supplement processes like Datasheet for Datasets and Data Statements for NLP \cite{Gebru2021DatasheetsFD, bender-friedman-2018-data}, which enable comprehensive documentation for benchmarks, but do not ask benchmark creators to reflect in a way that distinguishes between: 1) how they conceptualize a task (and how others may disagree with their conceptualization), and 2) how well the benchmark operationalizes a measurement model for model performance on their conceptualization of the task. This framework is also complementary to technical solutions (e.g., human-in-the-loop approaches) to resolving task ambiguity \cite{Tamkin2022TaskAI}.

We hope that this reflection will benefit the NLP community in the following ways:
\begin{itemize}[noitemsep,topsep=0pt,parsep=0pt,partopsep=0pt,leftmargin=*]
    \item \textbf{Reduce overhyping}: By being transparent about and defining the model capabilities that benchmarks are intended to assess, as well as documenting benchmark validity issues, benchmark creators will: 1) not misrepresent model capabilities, and 2) remind people to be careful about extrapolating benchmark performance results. 
    \item \textbf{Encourage reflexivity and engagement with the politics of benchmarks:} By clarifying how they conceptualize tasks and considering how others may disagree with their conceptualization, benchmark creators will: 1) assess how their social context and power influences task conceptualization and benchmark construction \cite{Collins2017BlackFT}, 2) reflect on which groups of people benchmarks represent, and 3) include people from diverse communities during benchmark construction towards alleviating disagreement. Towards considering historical and social context, we urge practitioners to not neutralize disagreements in conceptualization by valuing all ``sides'' equally, as this inevitably invalidates marginalized people's lived experiences and perpetuates the power relations in which benchmark construction participates \cite{Collins2017BlackFT, Denton2021WhoseGT}. In particular, the widespread adoption, presumed validity, and inertia of benchmarks influence the direction of NLP, shaping funding landscapes and the domains in which NLP systems are deployed \cite{BliliHamelin2022MakingIE, Bommasani2022E4C}. As such, we encourage practitioners to prioritize the perspectives of marginalized people.
    \item \textbf{Provide actionable insights to address benchmark validity issues:} Distinguishing between conceptualization and operationalization disagreements in a benchmark will better enable the creators of the benchmark, as well as creators of future benchmarks, to address benchmark validity issues. For example, to address perceptions that a benchmark does not exhaustively assess whether models can ``handle real-world phenomena,'' benchmark creators can decide if this is a conceptualization disagreement (e.g., ``real-world'' is too open-ended, in which case creators should clearly explain which domains they foreground in their conceptualization of ``real-world'') or operationalization disagreement (e.g., acquiring real-world data is difficult.) 
\end{itemize}

\section{Conclusion}

 We develop a taxonomy of disagreement (based on measurement modeling) which distinguishes between how tasks are conceptualized and how measurements of model performance are operationalized. To provide evidence for our taxonomy, we conduct a survey of practitioners and meta-analysis of relevant literature. Based on our taxonomy, we propose a framework for the creation of benchmarks and the documentation of their limitations. Future work includes studying task conceptualization via benchmark inter-annotator disagreement.
 
\clearpage

\section*{Limitations}
 
\paragraph{Survey limitations} Our survey sample size over-represents English-speaking NLP practitioners, and likely practitioners from the United States. While we would like to study the demographic skews in our sample (e.g., seniority) and its implications for the results in our paper, we could not collect demographic data due to privacy concerns. Nevertheless, our results still highlight that even within skewed samples, there exists weak agreement on how tasks are conceptualized.
Additionally, we assume that survey participants do not base their perceptions of task conceptualization on surface characteristics of tasks, or task ethos (e.g., task longevity, task popularity, rhetoric associated with the task). Furthermore, while we provide some justification for the 6-point scale in Appendix~\ref{sec:participant_guidance}, the scale is not optimal, as not many participant judgments are below 4; we had not run a similar survey previously, nor did our pilot responses indicate that many judgments would be $\geq 4$. Finally, while we would like to provide a qualitative analysis of participants’ free responses, the majority of participants did not answer the ``Additional Thoughts'' questions.

\paragraph{Meta-analysis limitations} We largely focus on static textual single-task English-language benchmarks. Furthermore, we assume that the capabilities stated by authors generally represent the primary capabilities that they believe the task involves; however, authors may refrain from including particular information due to space limits or reviewing incentives.

\paragraph{Framework limitations} While our proposed framework for creating benchmarks has not been explicitly tested, we have confidence in its efficacy as it was borne out of our systematic analysis of NLP practitioners, literature, and benchmarks. We ultimately wish to implement the framework, but doing so is beyond the scope of this paper (whose primary focus is a systematic perspective on disagreements on evaluative practices in NLP), and leave it to future work.

 \section*{Ethics and Broader Impact} 

We obtained informed consent from all survey participants, and the survey was IRB-approved. In administering the survey, we did not collect any personally identifiable information that could be traced back to participants' responses, and we transparently communicated our data privacy, usage, and retention policies (refer to \Cref{sec:survey_consent_form}). Furthermore, we shared our survey with artificial intelligence affinity groups to increase the diversity of our sample. We detail our participant recruitment protocol and IRB approval in \Cref{sec:survey_participant_recruitment_irb}. Additionally, in our paper, we discuss our taxonomy and benchmark documentation guidelines in the context of scientific accountability, power relations, and path dependence in NLP.

\section*{Acknowledgements}
We thank Alexander M. Hoyle and Eve Fleisig for their feedback on early versions of our survey. We further thank Li Lucy, Kai-Wei Chang, Jieyu Zhao, Swabha Swayamdipta, and the reviewers for their feedback on the writing of this paper.

\clearpage

\bibliography{anthology,custom}
\bibliographystyle{acl_natbib}

\newpage
\appendix

\onecolumn

\begin{center}
    \Large \textbf{Appendix}
\end{center}

\textbf{\large{Contents in Appendices:}}
\begin{itemize}
    \item In Appendix~\ref{sec:task_disambiguation}, we disambiguate the definition of ``benchmarks'' and ``tasks.''
    \item In Appendix~\ref{sec:issue_threaten_validity}, we provide examples of issues that threaten the validity of NLP benchmarks.
    \item In Appendix~\ref{sec:disagreements_supp}, we supply an extended discussion of how to distinguish between conceptualization and operationalization disagreements.
    \item In Appendix~\ref{sec:survey_task_selection}, we explain how we selected the tasks in our survey.
    \item In Appendix~\ref{sec:participant_guidance}, we detail the guidance we provided to survey participants.
    \item In Appendix~\ref{sec:survey_quality_control}, we detail the quality control protocol we followed for our survey.
    \item In Appendix~\ref{sec:survey_participant_recruitment_irb}, we detail how we recruited and compensated survey participants.
    \item In Appendix~\ref{sec:survey_questions}, we provide the full script of our survey, including the consent form.
    \item In Appendix~\ref{sec:comprehensive_survey_result}, we provide plots that summarize the responses to our survey questions.
    \item In Appendix~\ref{sec:model_capabilities}, we provide our qualitative analyses of the model capabilities that papers claim NLP benchmarks assess.
    \item In Appendix~\ref{sec:essentially_contested_examples}, we offer additional examples of common essentially contested constructs in NLP.
    \item In Appendix~\ref{sec:additional_disagreements}, we offer additional examples of conceptualization disagreements for different NLP tasks.
\end{itemize}

\clearpage
\section{Disambiguating Benchmarks and Tasks}
\label{sec:task_disambiguation}

\begin{itemize}
\item \textbf{Benchmarks:} We refer to benchmarks for a specific NLP task rather than a benchmark suite \cite{Dehghani2021TheBL}. We further only consider benchmarks for evaluation and do not make assumptions about how models are trained.
\item \textbf{Tasks:} In NLP, ``task'' has been used to refer to a ``format'' or ``language-related skill'' \cite{Gardner2019QuestionAI}. A format is typically a behavior specification, including a ``way of posing a particular problem to a machine'' along with what is expected as output  \cite{Bowman2021WhatWI}. Consider summarization, which can vary in format: given a long passage of text, extractive summarization is about directly copying the most important spans from the passage, while abstractive summarization permits the generation of new sentences \cite{Narayan2018DontGM}. Some formats may be more amenable to certain real-world use cases or domains (e.g., clinical text, legal documents) than others. However, these various formats often capture a common language-related skill: capturing the main points from a longer passage of text using a few statements. Formats may capture the language-related skill underlying the task to varying degrees.

In this paper, we consider tasks that the NLP community has largely decided form a category (e.g., coreference resolution, question answering). These tasks can refer to a ``format,'' ``language-related skill,'' or both, and often have benchmarks specifically dedicated to them. Tasks may also overlap in ``format'' or ``language-related skill.'' For example, many consider the Winograd Schema Challenge to fall under the task of commonsense reasoning \cite{Levesque2011TheWS}, but the benchmark also assesses the ability of an NLP model to perform coreference resolution.  Tasks also ``can exist at varying granularities'' \cite{liao2021are}.
\end{itemize}

\newpage

\section{Issues that Threaten the Validity of NLP Benchmarks}
\label{sec:issue_threaten_validity}

\begin{table*}[!ht]
    \centering
    \resizebox{0.75\textwidth}{!}{\begin{tabular}{|p{0.3\textwidth} | p{0.6\textwidth}|}
        \toprule
        \multicolumn{1}{|c|}{\textbf{Benchmark issue}} & \multicolumn{1}{|c|}{\textbf{Prior research}} \\ \midrule
        data noise and errors & \cite[\textit{inter alia}]{schlegel-etal-2020-framework, Blodgett2021StereotypingNS, Northcutt2021PervasiveLE, Dziri2022OnTO} \\
        \midrule
        superficial cues (e.g., annotation artifacts) in the data &  \cite[\textit{inter alia}]{schwartz-etal-2017-effect, gururangan-etal-2018-annotation, poliak-etal-2018-hypothesis, Kaushik2018HowMR, McCoy2019RightFT, Kavumba2019WhenCP, Niven2019ProbingNN, Si2019WhatDB, Ramponi2022FeaturesOS, Friedman2022FindingDS} \\
        \midrule
        inherent annotator disagreement & \cite[\textit{inter alia}]{ovesdotter-alm-2011-subjective, plank-etal-2014-linguistically, pavlick-kwiatkowski-2019-inherent, Nie2020WhatCW, basile-etal-2021-need, Davani2022DealingWD, wong2022are, Sap2022AnnotatorsWA, Kanclerz2022What, Jiang2022InvestigatingRF} \\
        \midrule
        poor linguistic diversity & \cite[\textit{inter alia}]{Hossain2020AnAO, Hossain2022AnAO, Parmar2022DontBT, Selvam2022TheTW, Seshadri2022QuantifyingSB} \\
        \midrule
        task format unsuitability & \cite[\textit{inter alia}]{Kaushik2018HowMR, Chen2019UnderstandingDD} \\
        \midrule
        insufficiently fine-grained evaluation &  \cite[\textit{inter alia}]{lalor-etal-2016-building, rodriguez-etal-2021-evaluation, zhong-etal-2021-larger} \\
        \midrule
        poorly aligned metrics & \cite[\textit{inter alia}]{Wagstaff2012MachineLT, ethayarajh-jurafsky-2020-utility, marie-etal-2021-scientific, Deutsch2022ReExaminingSC, Moghe2022ExtrinsicEO} \\
         \bottomrule
    \end{tabular}}
    \caption{Prior research has surfaced issues with NLP benchmarks that call into question their validity as measurements of model performance.}
    \label{tab:benchmark_issues}
\end{table*}
\FloatBarrier

\newpage

\section{Extended Discussion of Conceptualization and Operationalization Disagreements}
\label{sec:disagreements_supp}

There often exists a blurry line between conceptualization and operationalization disagreements. This is because it can be difficult to ascertain that everyone in $P_{B_\tau}$ truly conceptualizes an aspect of $\tau$ in the same way. As such, every practitioner could conceivably conceptualize $\tau$ differently. However, \citet{Palomaki2018} argue that, while tasks are often ``inherently subjective,'' there exists ``acceptable variation'' in task conceptualization (e.g., in the case of $y_\tau$, ``there may be divergent annotations that are truly of unacceptable quality'').

Moreover, it is often challenging to impute how the creators of $B_\tau$ conceptualize $\tau$ solely from their stated goals (e.g., in the paper that proposes $B_\tau$), due to incomplete statements of $C_\tau$, $M_\tau$, and $\neg M_\tau$; ambiguous specifications of $y_\tau$; and unclear explanations (if any) of how the creators understand $E_\tau$ \cite{Jiang2022InvestigatingRF, Tamkin2022TaskAI}. However, distinguishing between conceptualization and operationalization disagreements is critical to contextualize progress in NLP.
As such, we argue for benchmark documentation practices wherein the creators of $B_\tau$ clearly and comprehensively delineate their conceptualization of $\tau$ (\Cref{sec:documentation}).

\newpage

\section{Survey Task Selection}
\label{sec:survey_task_selection}

\begin{itemize}
\item \textbf{Task selection:} We first sourced well-recognized tasks to potentially include in our survey from a variety of sources including the AllenNLP demo\footnote{\url{https://demo.allennlp.org/}} \cite{Gardner2018AllenNLPAD}, NLP-Progress\footnote{\url{http://nlpprogress.com/}}, Papers With Code\footnote{\url{https://paperswithcode.com/}}, and popular benchmark suites such as GLUE \cite{Wang2018GLUEAM}, SuperGLUE \cite{Wang2019SuperGLUEAS}, and GEM \cite{gehrmann-etal-2021-gem}. To keep our survey at a reasonable length, we shortlisted tasks that we perceive, based on a cursory literature review (described below), to fall on a spectrum with respect to factors \textbf{(1)} and \textbf{(2)}. Our survey results suggest that our perceptions generally agree with those of the broader community.

\item \textbf{Literature selection:} To analyze factors \textbf{(1)} and \textbf{(2)} for various tasks, we identified relevant literature by inputting the search queries  ``[\textsc{Task}] survey'' and ``[\textsc{Task}] challenges'' into the Semantic Scholar search engine\footnote{\url{https://www.semanticscholar.org/}}. We considered the top 50 returned papers (sorted by ``Relevance''), for each also considering the papers it cites and that cite it. 
\end{itemize}

\section{Participant Guidance}
\label{sec:participant_guidance}
To ground participants' responses, we disambiguate ``task'' (\Cref{sec:task_disambiguation}) and provide task definitions (\Cref{sec:survey_questions}). However, because we are interested in participants' perceptions, we purposely do not prescribe definitions for terms like ``performance,'' ``progress,'' and ``state-of-the-art.'' For each task, we also first ask participants to list associated benchmark datasets and metrics with which they are familiar to further ground their responses to questions about general benchmark quality. Moreover, for all questions where participants are asked to rate their perception, we provide a scale that ranges from 1 to 6 with articulations of what 1 and 6 mean in the context of the question. We do this to: a) capture the distribution of participants' responses with sufficient granularity, b) impel participants to lean towards one side of the scale, and c) improve the consistency of how participants interpret answer choices. Finally, we do not specify language(s) for any tasks, including machine translation.

\clearpage

\section{Survey Quality Control}
\label{sec:survey_quality_control}
Before releasing our survey, we piloted it with a few industry practitioners ($N = 4$) in order to identify potential problems with the clarity of our questions. We further provided participants with the opportunity to optionally justify their responses or indicate disagreement or a lack of clarity with any definitions or questions. We intentionally included a few free-response questions (e.g., description of their NLP work) to deter and remove spammers from our sample. After filtering out spammers, we ultimately had $N = 46$ responses.

\section{Survey Participant Recruitment and IRB}
\label{sec:survey_participant_recruitment_irb}
We recruited survey participants who identify as NLP practitioners by sharing our survey as a Microsoft form on Twitter, NLP-focused Slack workspaces, and mailing lists or Slack channels for artificial intelligence (AI) affinity groups like Queer in AI \cite{jethwani2022}, Widening NLP.\footnote{
\url{https://www.winlp.org/}}, and Women in Machine Learning\footnote{\url{https://wimlworkshop.org/}}. We additionally shared the survey at a tech company via internal NLP mailing lists and an internal communication platform. In all cases, we requested participants to share the survey with other relevant groups in order to perform snowball sampling \cite{Naderifar2017SnowballSA}. Survey participants could optionally enter a raffle to win one of ten \$50 Amazon gift cards or virtual visa cards (depending on location of residence\footnote{While practitioners from anywhere in the world were welcome to participate in our survey, participants from certain countries were not eligible to enter the raffle due to local laws or gift card supplier rules.}). Given that many participants receive no money, the raffle is not adequate payment in all countries of residence. We obtained informed consent (refer to \Cref{sec:survey_consent_form}) from all survey participants, and the survey was IRB-approved.

\newpage

\section{Survey Questions and Responses}
\label{sec:survey_questions}

\noindent * Indicates required questions. \\

\noindent \textbf{Perceptions of Conceptualization and Evaluation of Natural Language Processing (NLP) Tasks} \\

\noindent We are interested in understanding how NLP practitioners and researchers perceive how well conceptualized and evaluated NLP tasks are. We hope that by understanding such perceptions, we will be able to better unpack validity issues with existing NLP benchmarks. \\ 

\noindent \textbf{What is an NLP task?} In NLP, ``task'' has been used to refer to a ``format'' or ``language-related skill'' \cite{Bowman2021WhatWI}. A format is typically a behavior specification, including a ``way of posing a particular problem to a machine'' along with what is expected as output \cite{Gardner2019QuestionAI}. Consider summarization, which can vary in format: given a long passage of text, extractive summarization is about directly copying the most important spans from the passage, while abstractive summarization permits the generation of new sentences. Some formats may be more amenable to certain real-world use cases or domains (e.g., clinical text, legal documents, etc.) than others. However, these various formats are often designed to capture a common language-related skill: capturing the main points from a longer passage of text using a few statements. Different formats may capture the language-related skill underlying the task to varying degrees. \\

\noindent If you have any questions, please feel free to contact us at: [REDACTED FOR ANONYMITY]. 

\subsection{Consent Form}
\label{sec:survey_consent_form}

\subsubsection{Introduction}
Thank you for taking the time to consider volunteering in a [REDACTED FOR ANONYMITY] research project. This form explains what would happen if you join this research project. Please read it carefully and take as much time as you need. Ask the study team about anything that is not clear. You can ask questions about the study any time. Participation in this study is voluntary and you will not be penalized if you decide not to take part in the study or if you quit the study later. \\

\noindent Project Name: Perception of Formulation and Evaluation of Natural Language Processing (NLP) Tasks \\
\noindent Principal Investigator: [REDACTED FOR ANONYMITY] \\
\noindent Other Investigators: [REDACTED FOR ANONYMITY]

\subsubsection{Purpose}
The purpose of this project is to audit popular benchmark datasets that are commonly used to assess natural language models’ performance on a range of NLP tasks, with a focus on issues related to validity. To select a subset of NLP tasks and associated datasets for our study, we would like to run an online survey of individuals who are working on these tasks and/or are familiar with evaluating NLP models to collect their opinions on the ambiguity, simplicity, and popularity of NLP tasks, as well as of the perceived quality of benchmark datasets and metrics associated with each task.

\subsubsection{Procedures}

During this project, you will complete a ~20-25-minute MS forms survey. \\

\noindent [REDACTED FOR ANONYMITY] may document and collect information about your participation through the answers you provide in the forms. No third parties will be involved in the transcription, processing, or analysis of the data. Approximately 100 participants will be involved in this study. You can copy or print this consent form for your own records, or you can email us at [REDACTED FOR ANONYMITY] for a copy of this form.

\subsubsection{Study Information and Confidentiality}

\noindent [REDACTED FOR ANONYMITY] is ultimately responsible for determining the purposes and uses of your study information. \\

\noindent \textbf{How we use study information.} The study information and other data collected during this project will be used primarily to perform research for purposes described in the introduction above.  Such information and data, or the results of the research may eventually be used to develop and improve our commercial products, services or technologies. \\

\noindent \textbf{Personal information we collect.} During this project, if you choose to enter the sweepstakes and provide the required personal information, we will collect details such as first name, last name, email address, and country of residence. \\

\noindent \textbf{How we store and share your study information.} Your name and other personal information will not be on the study information we receive about you or from you; the personal information will be identified by a code (e.g., a key phrase you provide) and this personal information will be kept separate from your study information, in a secured, limited access location. We will use this code only to ensure that those signing up for the sweepstakes have answered the survey and are not spammers. If you chose not to enter the sweepstakes, no personally identifiable information will be collected about you. \\

\noindent Your study information will stored for a period of up to 18 months. \\

\noindent Aside from the researchers of this study, your study information may be shared with study team members outside of [REDACTED FOR ANONYMITY], applicable individuals within [REDACTED FOR ANONYMITY], but confidentiality will be maintained, as allowed by law. \\

\noindent \textbf{How you can access and control your personal information.} If you wish to review or copy any personal information you provided during the study, or if you want us to delete or correct any such data, email your request to the research team at: [REDACTED FOR ANONYMITY]. \\

\noindent For additional information or concerns about how [REDACTED FOR ANONYMITY] handles your personal information, please see the [REDACTED FOR ANONYMITY] Privacy Statement ([REDACTED FOR ANONYMITY]).

\subsubsection{Benefits and Risks}

\noindent \textbf{Benefits:} There are no direct benefits to you that might reasonably be expected as a result of being in this study. We seek to audit popular benchmark datasets that are commonly used to assess NLP model performance on a range of NLP tasks, particularly focusing on issues related to validity. In doing so, we will bring light to issues with current evaluation practices in NLP and their implications for the claims made about NLP model performance. We hope to publish a paper and develop guidance or tools for how practitioners could audit their benchmark datasets. \\

\noindent \textbf{Risks:} The risks of participating in this study are no greater than those encountered in everyday life. To help reduce such risks, all identifiers will be removed from the survey responses. The primary contact and investigator have completed IRB training, including safe data handling practices. We will update survey respondents with research outcomes.

\subsubsection{Future Use of Your Identifiable Information}

\noindent Identifiers might be removed from your identifiable private information, and after such removal, the information could be used for future research studies or distributed to another investigator for future research studies without your (or your legally authorized representative’s) additional informed consent.

\subsubsection{Payment for Your Participation}

At the end of the survey, you will be asked if you want to participate in a raffle for one of ten \$50 USD Amazon gift cards or equivalent virtual visa cards (depending on location of residence). Your odds of winning depend on the total number of participants but are no less than 1 in 10. Your data may be used to make new products, tests or findings. These may have value and may be developed and owned by [REDACTED FOR ANONYMITY] and/or others. If this happens, there are no plans to pay you. For Official Rules, see the PDF [REDACTED FOR ANONYMITY].

\subsubsection{Participation}

Taking part in research is always a choice. If you decide to be in the study, you can change your mind at any time without affecting any rights including payment to which you would otherwise be entitled. If you decide to withdraw, you should contact the person in charge of this study, and also inform that person if you would like your personal information removed as well. \\

\noindent [REDACTED FOR ANONYMITY] or the person in charge of this study may discontinue the study or your individual participation in the study at any time without your consent for reasons including:

\begin{itemize}
\item your failure to follow directions
\item it is discovered that you do not meet study requirements
\item it is in your best interest medically
\item the study is canceled
\item administrative reasons
\end{itemize}

\noindent If you leave the study, the study staff will still be able to use your information that they have already collected, however, you have the right to ask for it to be removed when you leave. Significant new findings that develop during the course of this study that might impact your willingness to be in this study will be given to you.

\subsubsection{Contact Information}

\noindent Should you have any questions concerning this project, or if you are injured as a result of being in this study, please contact: [REDACTED FOR ANONYMITY]. \\

\noindent Should you have any questions about your rights as a research subject, please contact the [REDACTED FOR ANONYMITY].

\subsubsection{Consent}

\noindent By completing this form, you confirm that the study was explained to you, you had a chance to ask questions before beginning the study, and all your questions were answered satisfactorily. At any time, you may ask other questions. By completing this form, you voluntarily consent to participate, and you do not give up any legal rights you have as a study participant. \\

\noindent Please confirm your consent by completing the bottom of this form. If you would like to keep a copy of this form, please print or save one. On behalf of [REDACTED FOR ANONYMITY], we thank you for your contribution and look forward to your research session. \\

\noindent [Q1] Do you understand and consent to these terms? *
\begin{itemize}[noitemsep,topsep=0pt,parsep=0pt,partopsep=0pt]
    \item[$\ocircle$] Yes
    \item[$\ocircle$] No \textcolor{red}{[if selected, survey branches to final section]}
\end{itemize}

\subsection{Background}

\noindent [Q2] Do you have any experience with NLP tasks? *
\begin{itemize}[noitemsep,topsep=0pt,parsep=0pt,partopsep=0pt]
    \item[$\ocircle$] Yes
    \item[$\ocircle$] No
\end{itemize}

\noindent [Q3] Briefly describe the type of NLP work that you do. *
\begin{itemize}[noitemsep,topsep=0pt,parsep=0pt,partopsep=0pt]
    \item Open text field
\end{itemize}

\noindent [Q4] Please select all the options that apply to you. *
\begin{itemize}[noitemsep,topsep=0pt,parsep=0pt,partopsep=0pt]
    \item[$\square$] I work on deployed systems
    \item[$\square$] I am an industry practitioner (not researcher)
    \item[$\square$] I am an industry researcher
    \item[$\square$] I am an academic researcher
\end{itemize}

\subsection{Perceived Performance}

\noindent [Q5] In general, how well do you think current state-of-the-art NLP models perform on the following tasks? Please select ``I don't know'' if you have never heard of the task or have little to no knowledge about it. * 

\noindent \textit{6 (high performance) means that you think current state-of-the-art NLP models tend to perform very well on this task, with little to no area for improvement. In contrast, 1 (low performance) means that current state-of-the-art models perform poorly on this task, including because the task is new or the task has been neglected by the community.}

\noindent 
\begin{tabular}{ccccccc}
1 & 2 & 3 & 4 & 5 & 6 & I don't know \\
$\ocircle$ & $\ocircle$ & $\ocircle$ & $\ocircle$ & $\ocircle$ & $\ocircle$ & $\ocircle$
\end{tabular}
\begin{itemize}[noitemsep,topsep=0pt,parsep=0pt,partopsep=0pt]
    \item Sentiment Analysis
    \item Natural Language Inference
    \item Question Answering
    \item Coreference Resolution
    \item Summarization
    \item Named-Entity Recognition
    \item Dependency Parsing
    \item Machine Translation
\end{itemize}

\subsection{NLP Task: Sentiment Analysis}
\label{sec:sentiment_analysis}

\noindent Given some input text, a model must correctly identify opinions, sentiments, and subjectivity in the text (read more: \url{https://en.wikipedia.org/wiki/Sentiment_analysis}) \cite{Pang2007OpinionMA}. \\

\noindent \textit{\textbf{Reminder:} ``Task'' has been used to refer to a ``format'' or ``language-related skill'' \cite{Bowman2021WhatWI}. A format is typically a behavior specification, including a ``way of posing a particular problem to a machine'' along with what is expected as output \cite{Gardner2019QuestionAI}. While NLP tasks can sometimes vary in format or domain, these various formats are often designed to capture a common language-related skill (e.g., summarization tasks try to capture the key points in a long passage of text using a few statements).  Different formats may capture the language-related skill underlying the task to varying degrees.} \\ 

\noindent [Q6] Are you familiar with this NLP task (including associated benchmark datasets and metrics)? *

\begin{itemize}[noitemsep,topsep=0pt,parsep=0pt,partopsep=0pt]
    \item[$\ocircle$] Yes, I am an expert (e.g., I have developed, deployed, researched, or evaluated NLP models on this task)
    \item[$\ocircle$] Yes, but I only have passing knowledge (e.g., I only have read, studied, or heard about this task)
    \item[$\ocircle$] No \textcolor{red}{[if selected, survey branches to next section]}
\end{itemize}

\noindent [Q7] \textbf{Task definition:} How well defined or conceptualized do you think this task is? * \\
\noindent \textit{6 (well defined) means that the task has an objective that is clearly and consistently articulated and understood by the NLP community. In contrast, 1 (poorly defined) means that the task has an objective that is understood differently from person to person in the community.} \\
\noindent 
\begin{tabular}{ccccccc}
1 & 2 & 3 & 4 & 5 & 6  \\
$\ocircle$ & $\ocircle$ & $\ocircle$ & $\ocircle$ & $\ocircle$ & $\ocircle$
\end{tabular}

\noindent [Q8] \textbf{Task instantiation:} In general, how well do you think common formats of this task capture the underlying language-related skill? * \\
\textit{6 (captures the skill well) means that common formats of this task perfectly capture the underlying language-related skill, while 1 (captures the skill poorly) means that common formats of this task do not capture the underlying language-related skill at all.} \\
\noindent 
\begin{tabular}{ccccccc}
1 & 2 & 3 & 4 & 5 & 6  \\
$\ocircle$ & $\ocircle$ & $\ocircle$ & $\ocircle$ & $\ocircle$ & $\ocircle$
\end{tabular}

\noindent [Q9] Write in any performance metrics for this task that you have experience with, if any. If none, please write ``N/A.'' * 
\begin{itemize}[noitemsep,topsep=0pt,parsep=0pt,partopsep=0pt]
    \item Open text field
\end{itemize}
\noindent [Q10] \textbf{Metrics quality:} In general, how well do you think common metrics (considering a broad range of metrics) capture NLP models' performance on this task? * \\
\textit{6 (captures performance well) means that metrics generally capture everything about performance on this task that we want it to capture, without capturing extraneous information. In contrast, 1 (captures performance poorly) means that metrics generally do not capture any valuable information about task performance or is highly influenced by extraneous signals.} \\
\noindent 
\begin{tabular}{ccccccc}
1 & 2 & 3 & 4 & 5 & 6  \\
$\ocircle$ & $\ocircle$ & $\ocircle$ & $\ocircle$ & $\ocircle$ & $\ocircle$
\end{tabular}

\noindent [Q11] Write in any benchmark datasets for this task that you have experience using, if any. If none, please write ``N/A.'' *
\begin{itemize}[noitemsep,topsep=0pt,parsep=0pt,partopsep=0pt]
    \item Open text field
\end{itemize}
\noindent [Q12] \textbf{Benchmark datasets quality:} In general, how would you assess the quality of benchmark datasets that are commonly used to evaluate NLP models on this task? * \\ 
\textit{6 (high dataset quality) means that the datasets generally are free of errors, and correctly and consistently capture the language-related skill underlying the task. In contrast, 1 (low dataset quality) means that the datasets generally contain significant errors or fail to capture the language-related skill underlying the task correctly and consistently.} \\
\noindent 
\begin{tabular}{ccccccc}
1 & 2 & 3 & 4 & 5 & 6  \\
$\ocircle$ & $\ocircle$ & $\ocircle$ & $\ocircle$ & $\ocircle$ & $\ocircle$
\end{tabular}

\noindent [Q13] \textbf{Current progress:} How close do you think current state-of-the-art NLP models are to learning the language-related skill underlying this task? *
\textit{6 (close) means that you think current state-of-the-art NLP models have successfully learned the language-related skill underlying this task. In contrast, 1 (not close) means that you think current state-of-the-art NLP models are still far from learning this skill.} \\
\noindent 
\begin{tabular}{ccccccc}
1 & 2 & 3 & 4 & 5 & 6  \\
$\ocircle$ & $\ocircle$ & $\ocircle$ & $\ocircle$ & $\ocircle$ & $\ocircle$
\end{tabular}

\noindent [Q14] \textbf{Potential progress:} How likely do you think NLP models are to ever learning the language-related skill underlying this task? \\ 
\textit{6 (highly likely) means that you think current state-of-the-art NLP models have learned or will surely learn the language-related skill underlying this task, while 1 (highly unlikely) means that you think NLP models will likely never learn this skill.} \\ 
\noindent 
\begin{tabular}{ccccccc}
1 & 2 & 3 & 4 & 5 & 6  \\
$\ocircle$ & $\ocircle$ & $\ocircle$ & $\ocircle$ & $\ocircle$ & $\ocircle$
\end{tabular}

\noindent [Q15] Do you have additional thoughts in response to the definitions and questions for this task? \\
\noindent \textit{This can include justifications of your responses or a lack of clarity on any definitions or questions. For example, do you agree with the task definition?}
\begin{itemize}[noitemsep,topsep=0pt,parsep=0pt,partopsep=0pt]
    \item Open text field
\end{itemize}

\subsection{NLP Task: Natural Language Inference}
Given a pair of input sentences, a model must correctly determine if the sentences satisfy a certain semantic relationship (e.g., textual entailment) (read more: \url{https://paperswithcode.com/task/natural-language-inference}) \cite{Storks2019RecentAI}. \\ 

\noindent Rest of section is same as \Cref{sec:sentiment_analysis}.

\subsection{NLP Task: Question Answering}
Given some knowledge source (e.g., a passage, image, knowledge base), a model must correctly answer given questions (read more: \url{https://en.wikipedia.org/wiki/Question_answering})  \cite{Gardner2019QuestionAI}. \\ 

\noindent Rest of section is same as \Cref{sec:sentiment_analysis}.

\subsection{NLP Task: Coreference Resolution}
Given some input text, a model must correctly identify expressions that refer to the same entity (read more: \url{https://en.wikipedia.org/wiki/Coreference#Coreference_resolution}) \cite{pradhan-etal-2011-conll}.  \\ 

\noindent Rest of section is same as \Cref{sec:sentiment_analysis}.

\subsection{NLP Task: Summarization}
Given some input text, a model must output a shorter summary that preserves key information from the input text (\url{read more: https://en.wikipedia.org/wiki/Automatic_summarization}) \cite{Allahyari2017TextST}.  \\ 

\noindent Rest of section is same as \Cref{sec:sentiment_analysis}.

\subsection{NLP Task: Named-Entity Recognition}
Given some input text, a model must correctly identify named entities (people, locations, organizations) in the text (read more: \url{https://en.wikipedia.org/wiki/Named-entity_recognition})  \cite{tjong-kim-sang-de-meulder-2003-introduction}.   \\ 

\noindent Rest of section is same as \Cref{sec:sentiment_analysis}.

\subsection{NLP Task: Dependency Parsing}
Given some input text, a model must correctly identify head words in the text and the dependent words which modify those heads (read more: \url{https://paperswithcode.com/task/dependency-parsing}) \cite{nivre-etal-2007-conll}. \\ 

\noindent Rest of section is same as \Cref{sec:sentiment_analysis}.

\subsection{NLP Task: Machine Translation}
Given some input content (e.g., text, video), a model must correctly translate the content from the source language to a target language (read more: \url{https://en.wikipedia.org/wiki/Machine_translation}) \cite{white-oconnell-1993-evaluation, yin-etal-2021-including}.  \\ 

\noindent Rest of section is same as \Cref{sec:sentiment_analysis}.

\subsection{Raffle Entry}

\noindent [Q86] [OPTIONAL] If you would like to enter the raffle drawing for one of the ten \$50 Amazon gift cards or equivalent virtual visa cards (depending on location of residence), for anonymity purposes, after submitting this form you will be provided with a link to another form to fill in your email address and enter the raffle. For this, \textbf{please also write down a key phrase here, which you will also be asked to re-enter on the raffle form}. We will only use this key phrase to validate that the raffle participants have completed the survey. Please don’t use a key phrase that is associated with any accounts. 
\begin{itemize}[noitemsep,topsep=0pt,parsep=0pt,partopsep=0pt]
    \item Open text field
\end{itemize}

\subsection{Feedback}

\noindent [Q87] Do you have any comments or feedback on the questions in this survey? \\
\noindent \textit{Please be mindful not to bring up any identifying or sensitive information about yourself or third-parties.}
\begin{itemize}[noitemsep,topsep=0pt,parsep=0pt,partopsep=0pt]
    \item Open text field
\end{itemize}

\newpage

\section{Comprehensive Survey Results}
\label{sec:comprehensive_survey_result}

\subsection{Perceived performance}

\begin{figure}[!ht]
    \centering
    \includegraphics[width=0.75\textwidth]{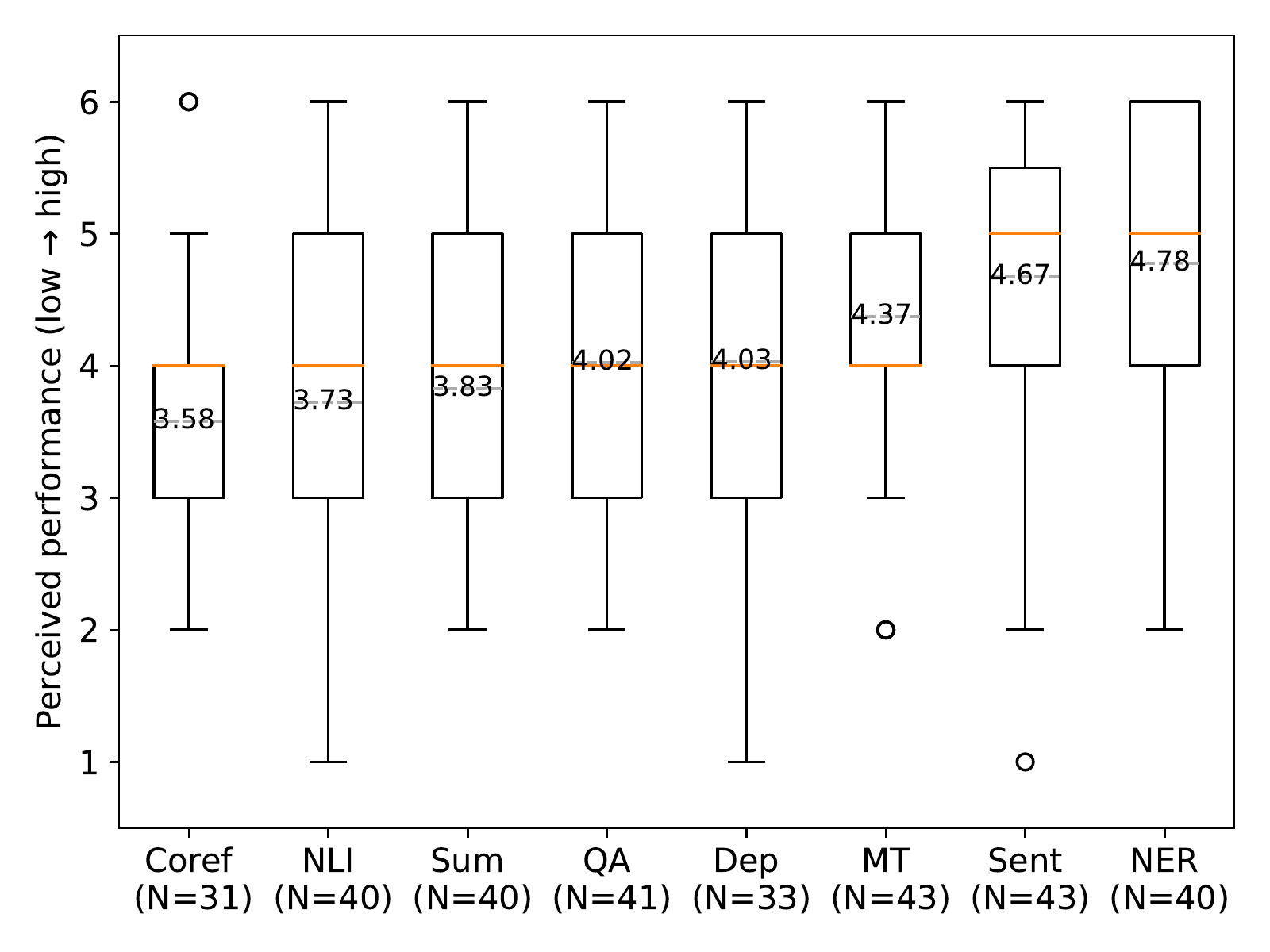}
    \caption{Perceived performance for all tasks.}
    \label{fig:performance}
\end{figure}
\FloatBarrier

\newpage

\subsection{Task definition}

\begin{figure}[!ht]
    \centering
    \begin{subfigure}[]{0.5\textwidth}
        \includegraphics[width=\textwidth]{survey_plots/Task_definition_ALL.pdf}
        \caption{Perceived clarity and consistency of task definition among all responding practitioners for each task.}
        \label{fig:task_definition_all}
    \end{subfigure}
    \begin{subfigure}[]{0.5\textwidth}
        \includegraphics[width=\textwidth]{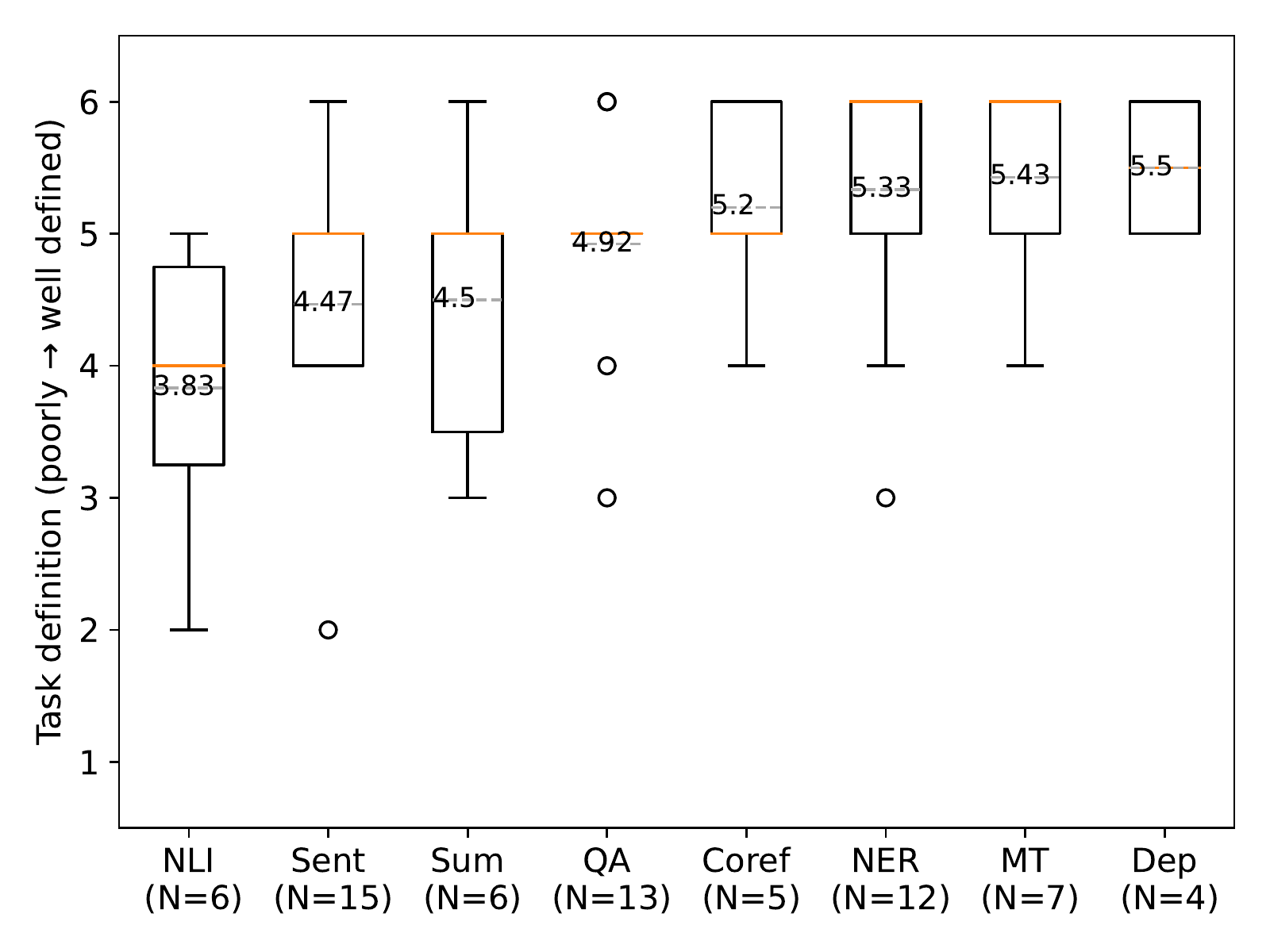}
        \caption{Perceived clarity and consistency of task definition among survey participants who consider themselves an ``expert'' at each task.}
        \label{fig:task_definition_expert}
    \end{subfigure}
    \begin{subfigure}[b]{0.5\textwidth}
        \includegraphics[width=\textwidth]{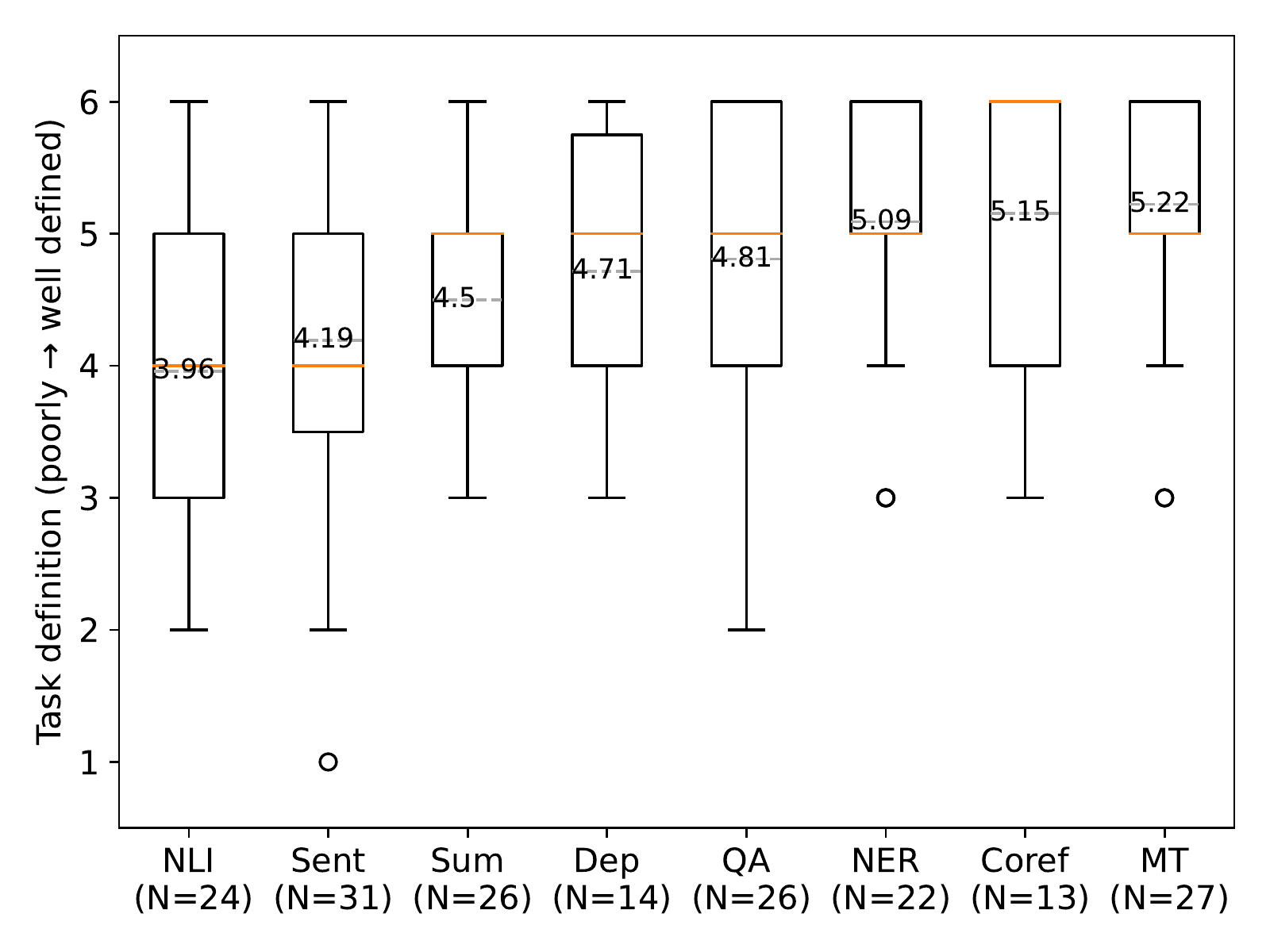}
        \caption{Perceived clarity and consistency of task definition among survey participants who consider themselves to have ``passing knowledge'' about each task.}
        \label{fig:task_definition_novice}
    \end{subfigure}
    \caption{}
    \label{fig:task_definition}
\end{figure}
\FloatBarrier

\newpage
\subsection{Task instantiation}

\begin{figure}[!ht]
    \centering
    \begin{subfigure}[]{0.5\textwidth}
        \includegraphics[width=\textwidth]{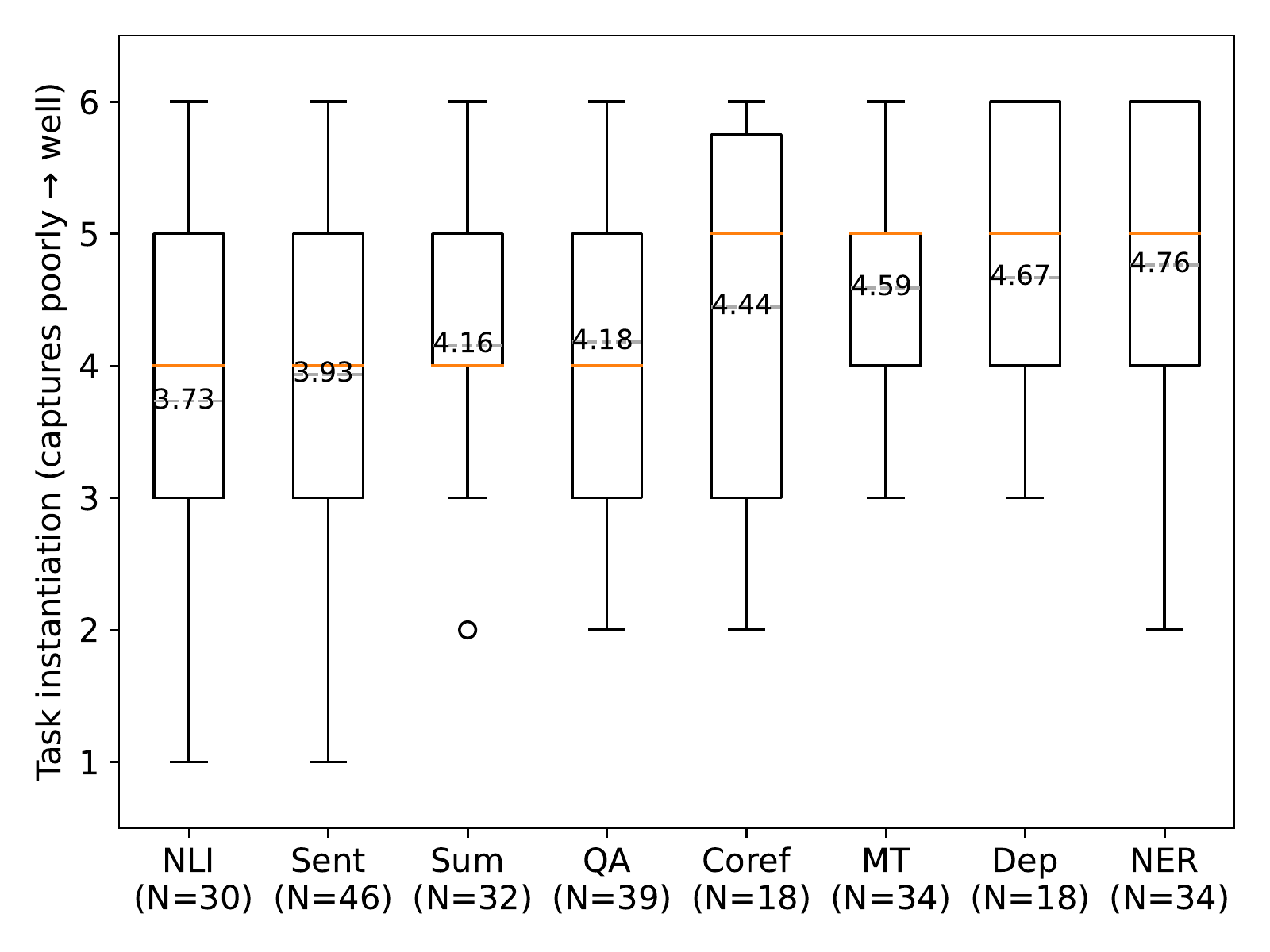}
        \caption{Perceived task instantiation quality among all responding practitioners for each task.}
        \label{fig:task_instantiation_all}
    \end{subfigure}
    \begin{subfigure}[]{0.5\textwidth}
        \includegraphics[width=\textwidth]{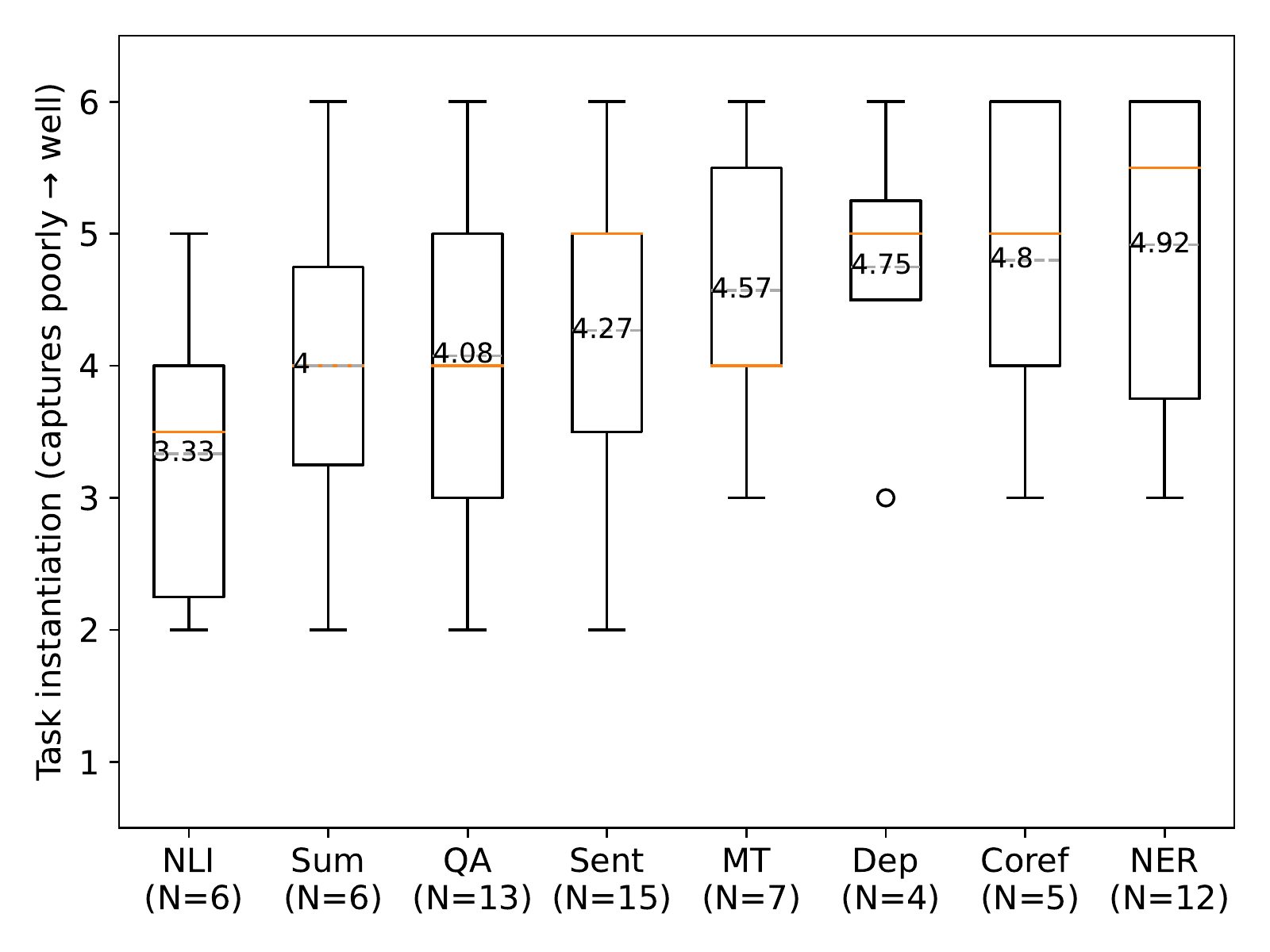}
        \caption{Perceived task instantiation quality among survey participants who consider themselves an ``expert'' at each task.}
        \label{fig:task_instantiation_expert}
    \end{subfigure}
    \begin{subfigure}[b]{0.5\textwidth}
        \includegraphics[width=\textwidth]{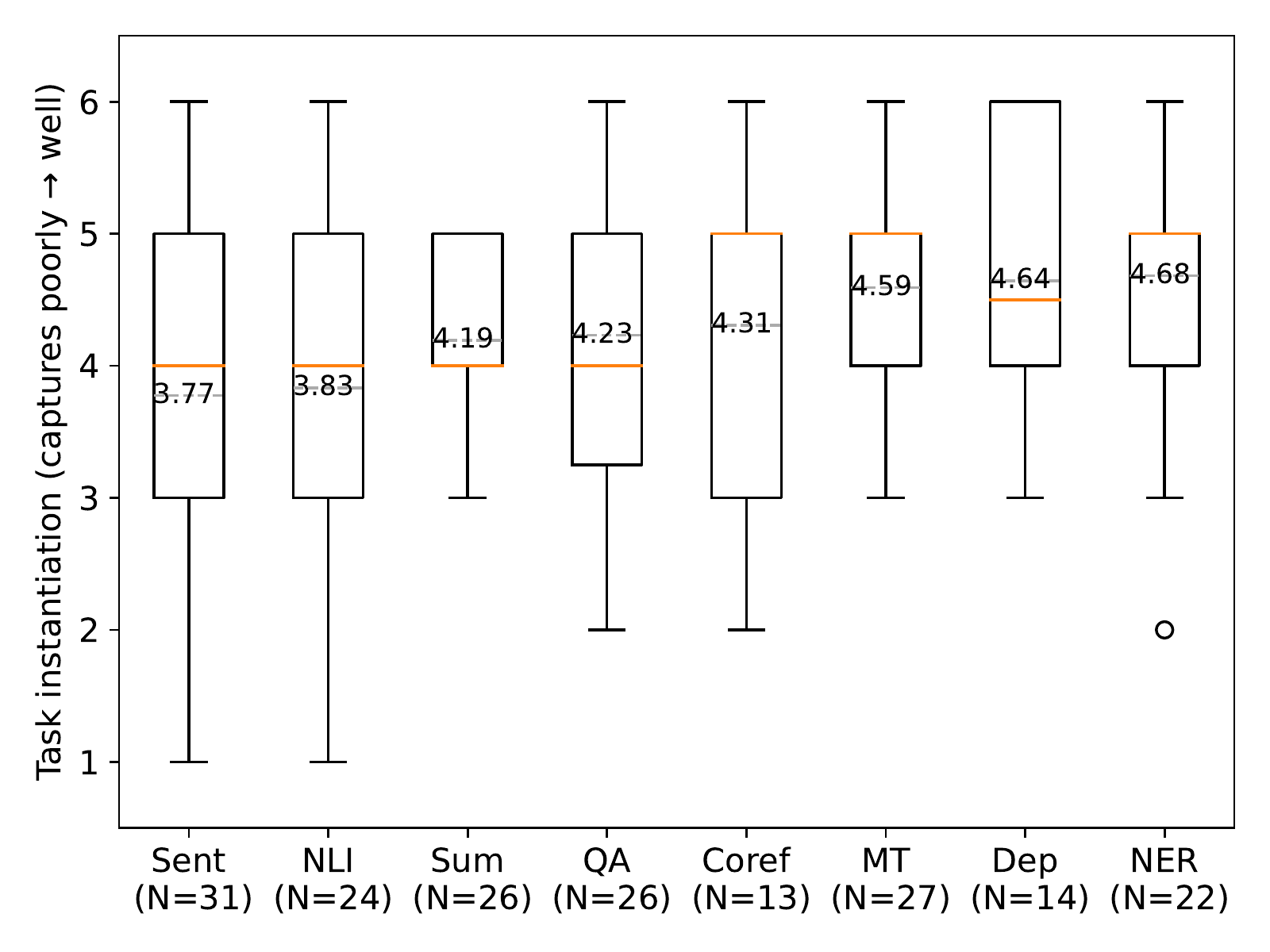}
        \caption{Perceived task instantiation quality among survey participants who consider themselves to have ``passing knowledge'' about each task.}
        \label{fig:task_instantiation_novice}
    \end{subfigure}
    \caption{}
    \label{fig:task_instantiation}
\end{figure}
\FloatBarrier

\newpage
\subsection{Metrics quality}
\begin{figure}[!ht]
    \centering
    \begin{subfigure}[]{0.5\textwidth}
        \includegraphics[width=\textwidth]{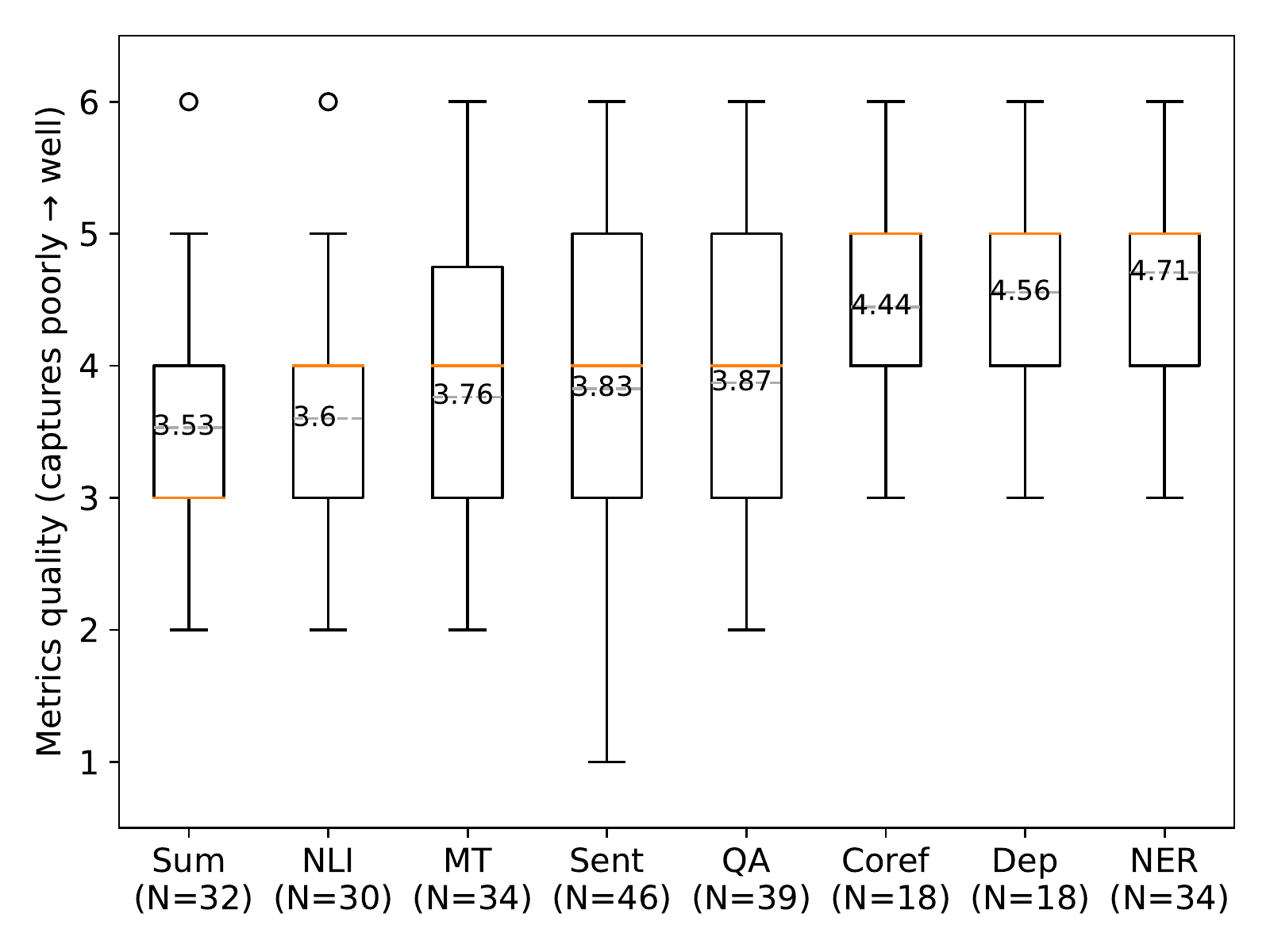}
        \caption{Perceived metrics quality among all responding practitioners for each task.}
        \label{fig:metrics_quality_all}
    \end{subfigure}
    \begin{subfigure}[]{0.5\textwidth}
        \includegraphics[width=\textwidth]{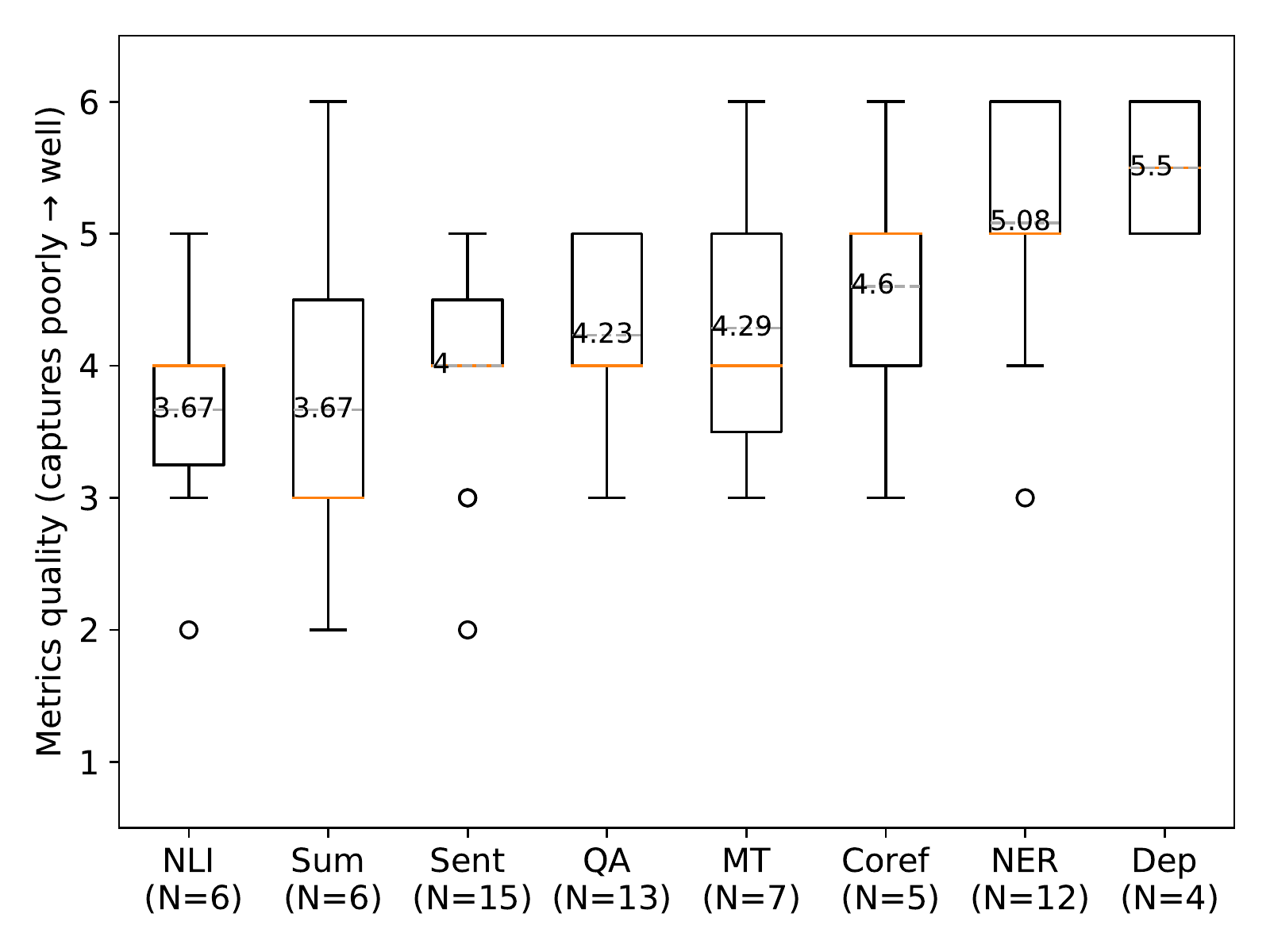}
        \caption{Perceived metrics quality among survey participants who consider themselves an ``expert'' at each task.}
        \label{fig:metrics_quality_expert}
    \end{subfigure}
    \begin{subfigure}[b]{0.5\textwidth}
        \includegraphics[width=\textwidth]{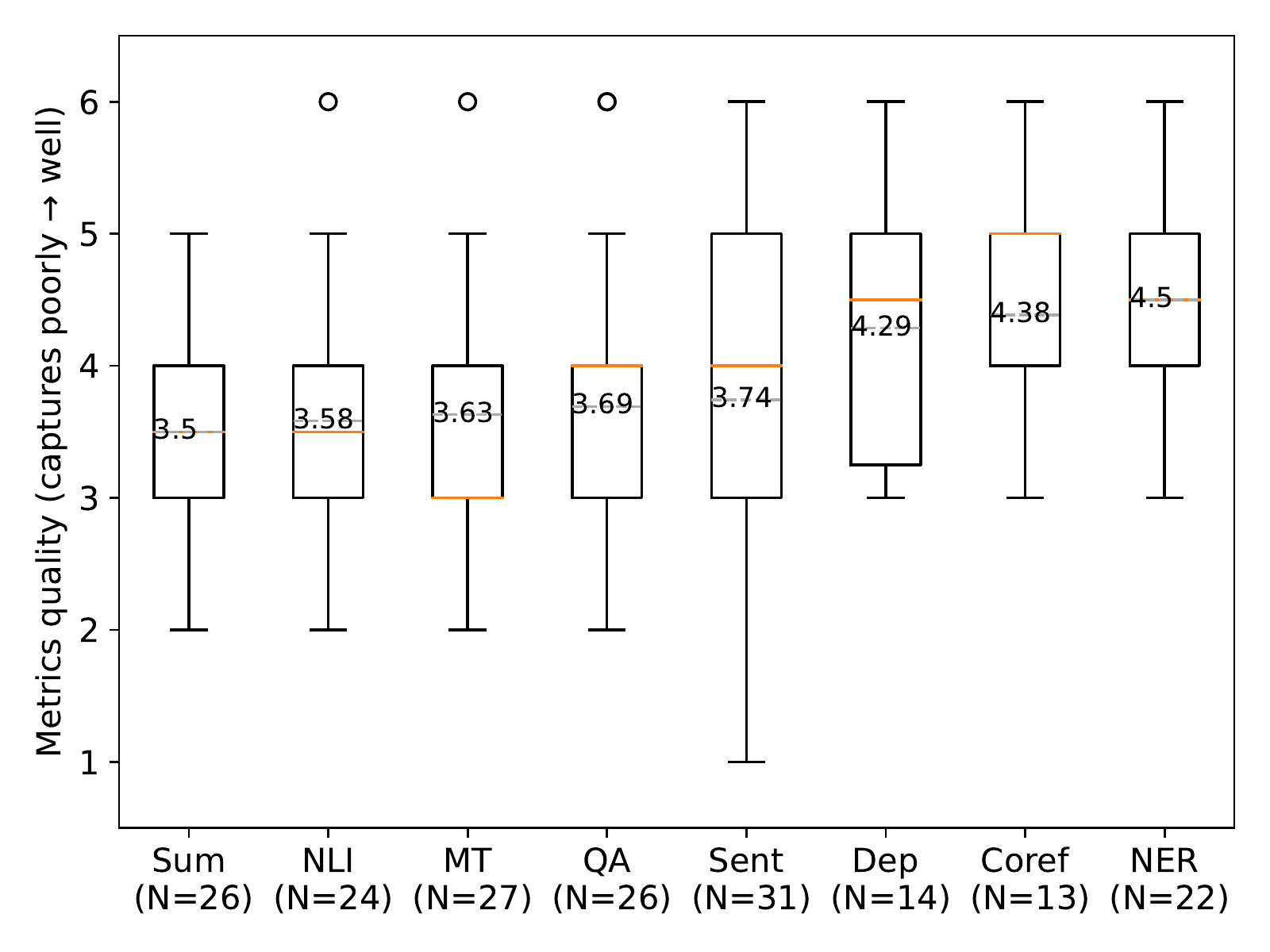}
        \caption{Perceived metrics quality among survey participants who consider themselves to have ``passing knowledge'' about each task.}
        \label{fig:metrics_quality_novice}
    \end{subfigure}
    \caption{}
    \label{fig:metrics_quality}
\end{figure}
\FloatBarrier

\newpage 

\subsection{Benchmark datasets quality}
\begin{figure}[!ht]
    \centering
    \begin{subfigure}[]{0.5\textwidth}
        \includegraphics[width=\textwidth]{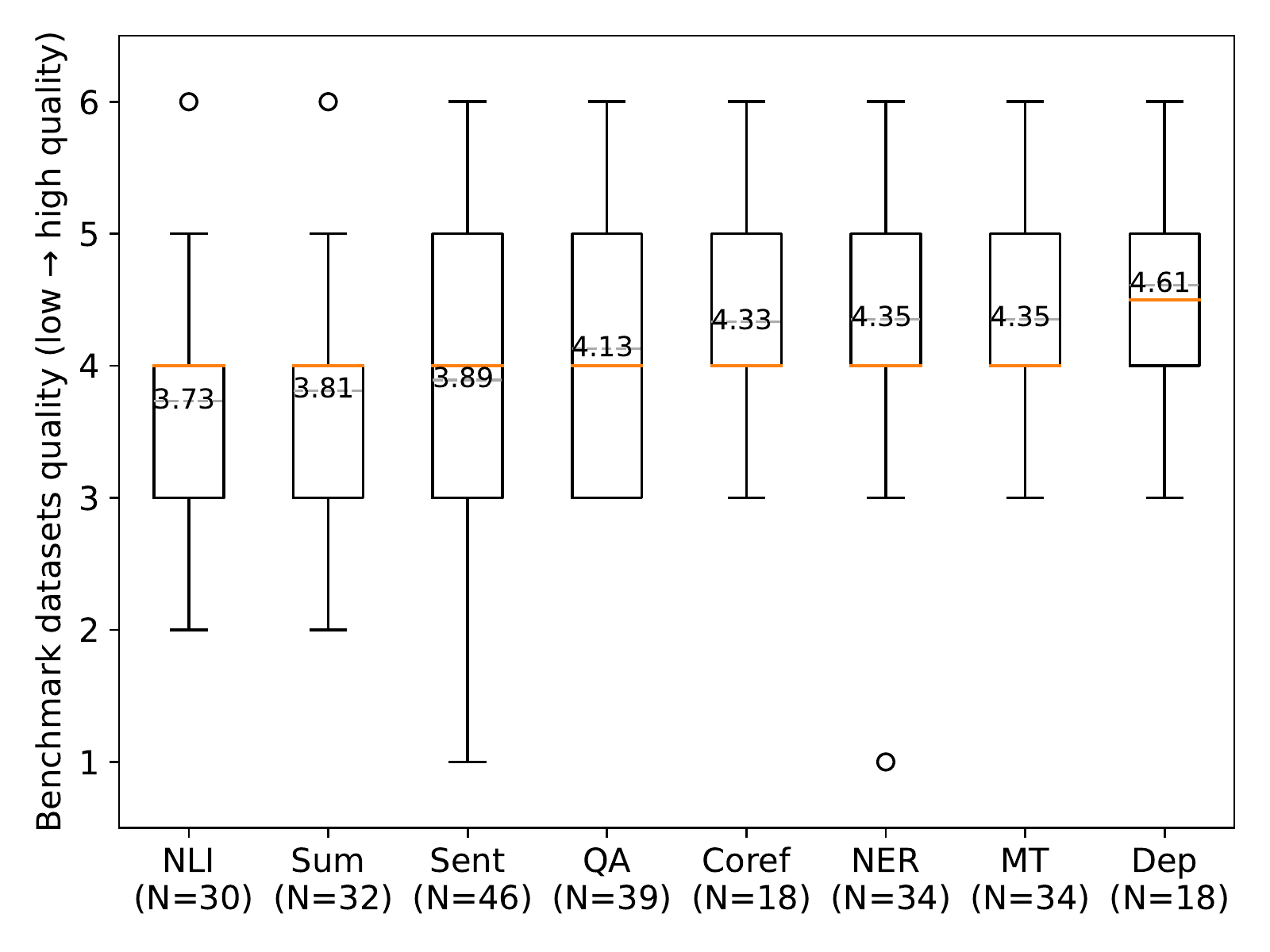}
        \caption{Perceived benchmark datasets quality among all responding practitioners for each task.}
        \label{fig:benchmarks_datasets_quality_all}
    \end{subfigure}
    \begin{subfigure}[]{0.5\textwidth}
        \includegraphics[width=\textwidth]{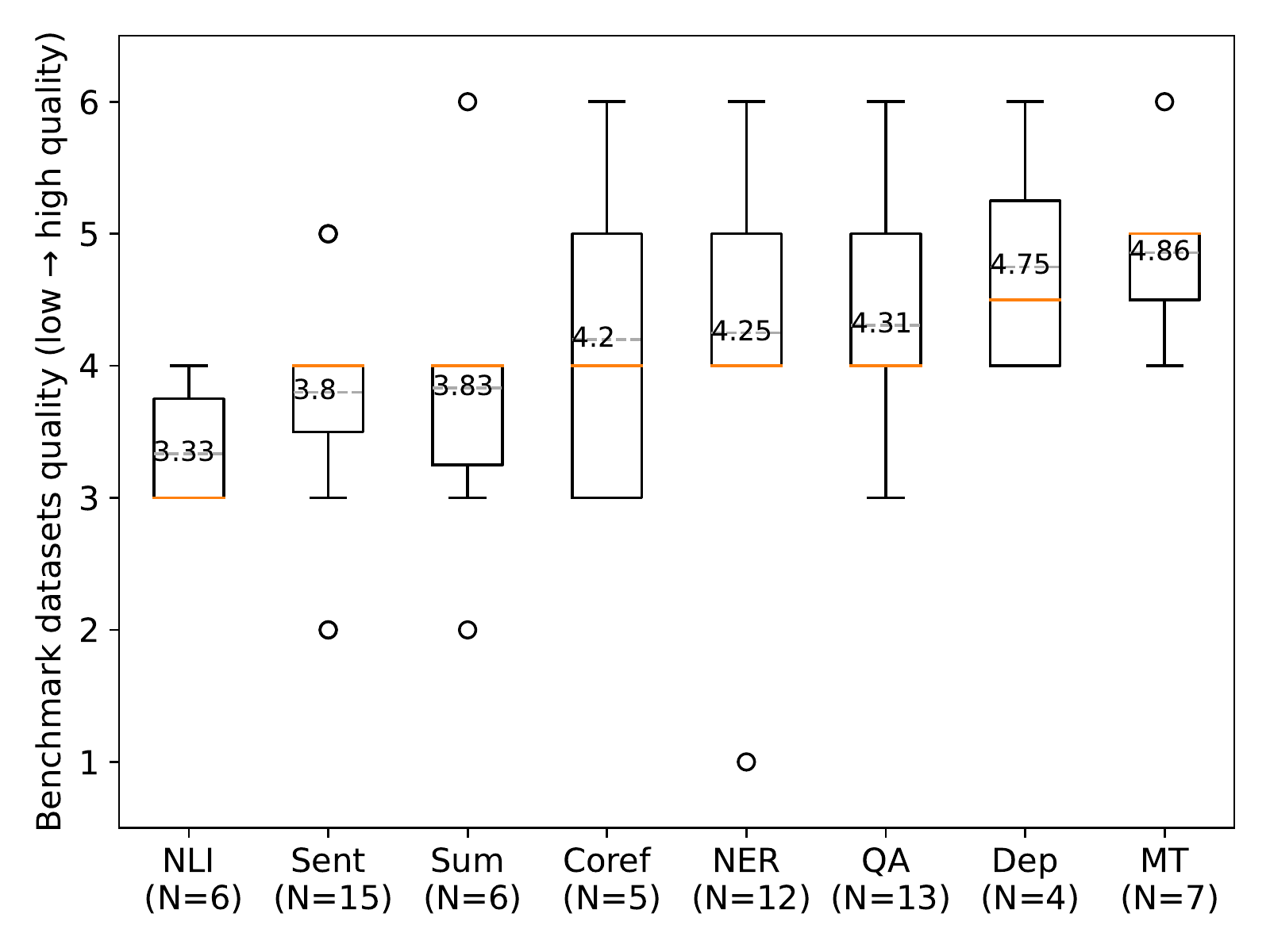}
        \caption{Perceived benchmark datasets quality among survey participants who consider themselves an ``expert'' at each task.}
        \label{fig:benchmarks_datasets_quality_expert}
    \end{subfigure}
    \begin{subfigure}[b]{0.5\textwidth}
        \includegraphics[width=\textwidth]{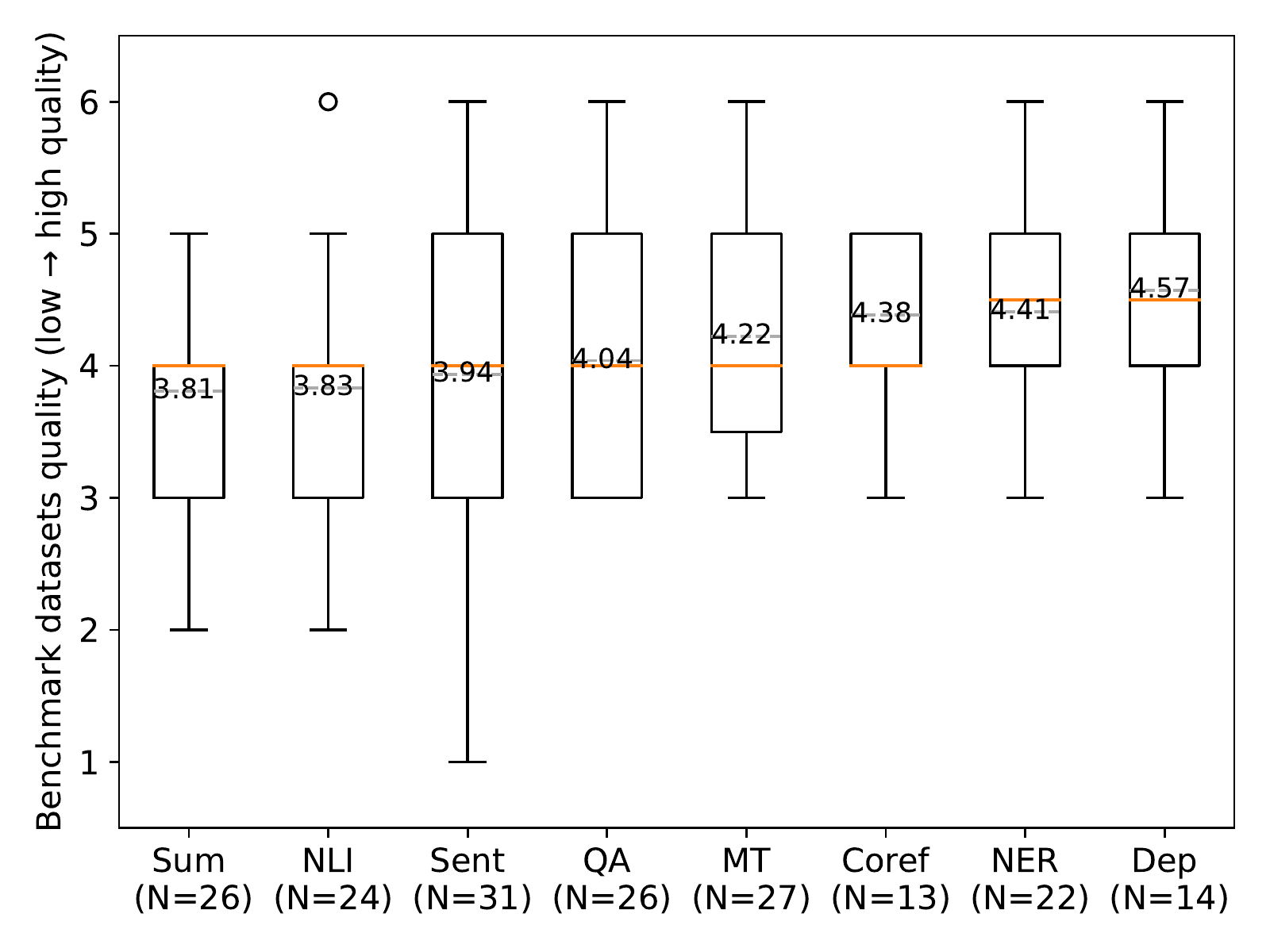}
        \caption{Perceived benchmark datasets quality among survey participants who consider themselves to have ``passing knowledge'' about each task.}
        \label{fig:benchmarks_datasets_quality_novice}
    \end{subfigure}
    \caption{}
    \label{fig:benchmarks_datasets_quality}
\end{figure}
\FloatBarrier

\newpage

\subsection{Current progress}
\begin{figure}[!ht]
    \centering
    \begin{subfigure}[]{0.5\textwidth}
        \includegraphics[width=\textwidth]{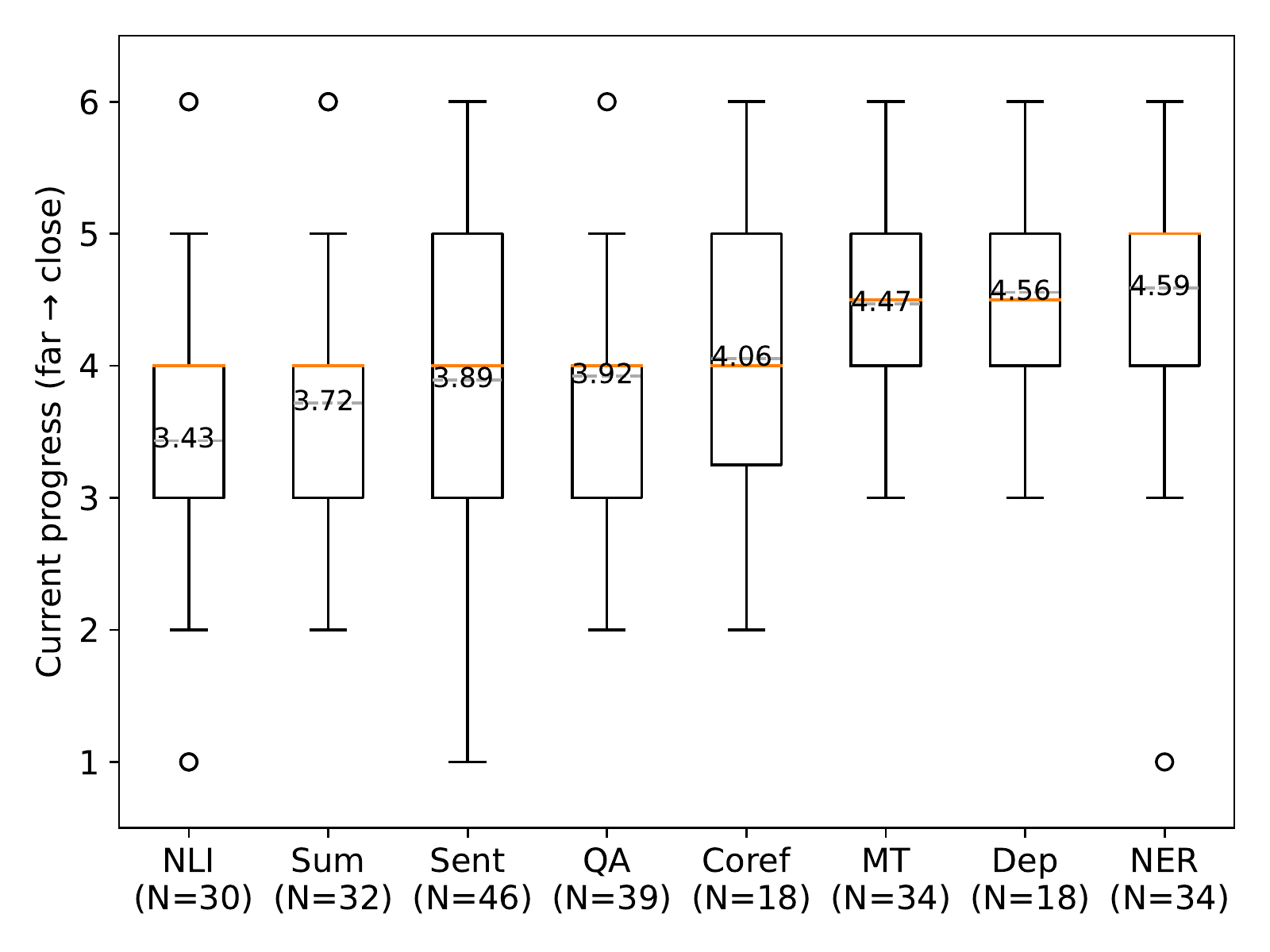}
        \caption{Perceived current progress among all responding practitioners for each task.}
        \label{fig:current_progress_all}
    \end{subfigure}
    \begin{subfigure}[]{0.5\textwidth}
        \includegraphics[width=\textwidth]{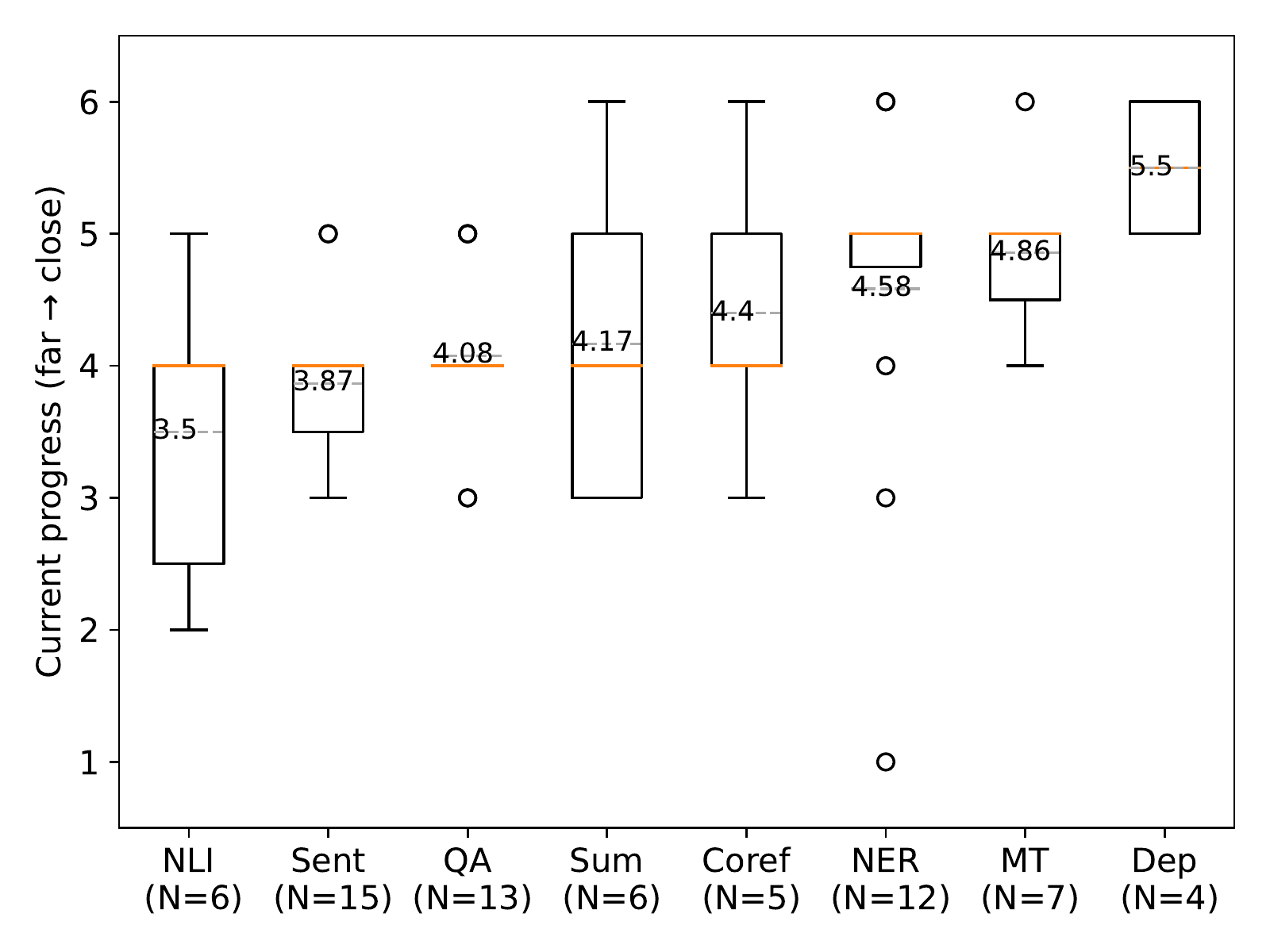}
        \caption{Perceived current progress among survey participants who consider themselves an ``expert'' at each task.}
        \label{fig:current_progress_expert}
    \end{subfigure}
    \begin{subfigure}[b]{0.5\textwidth}
        \includegraphics[width=\textwidth]{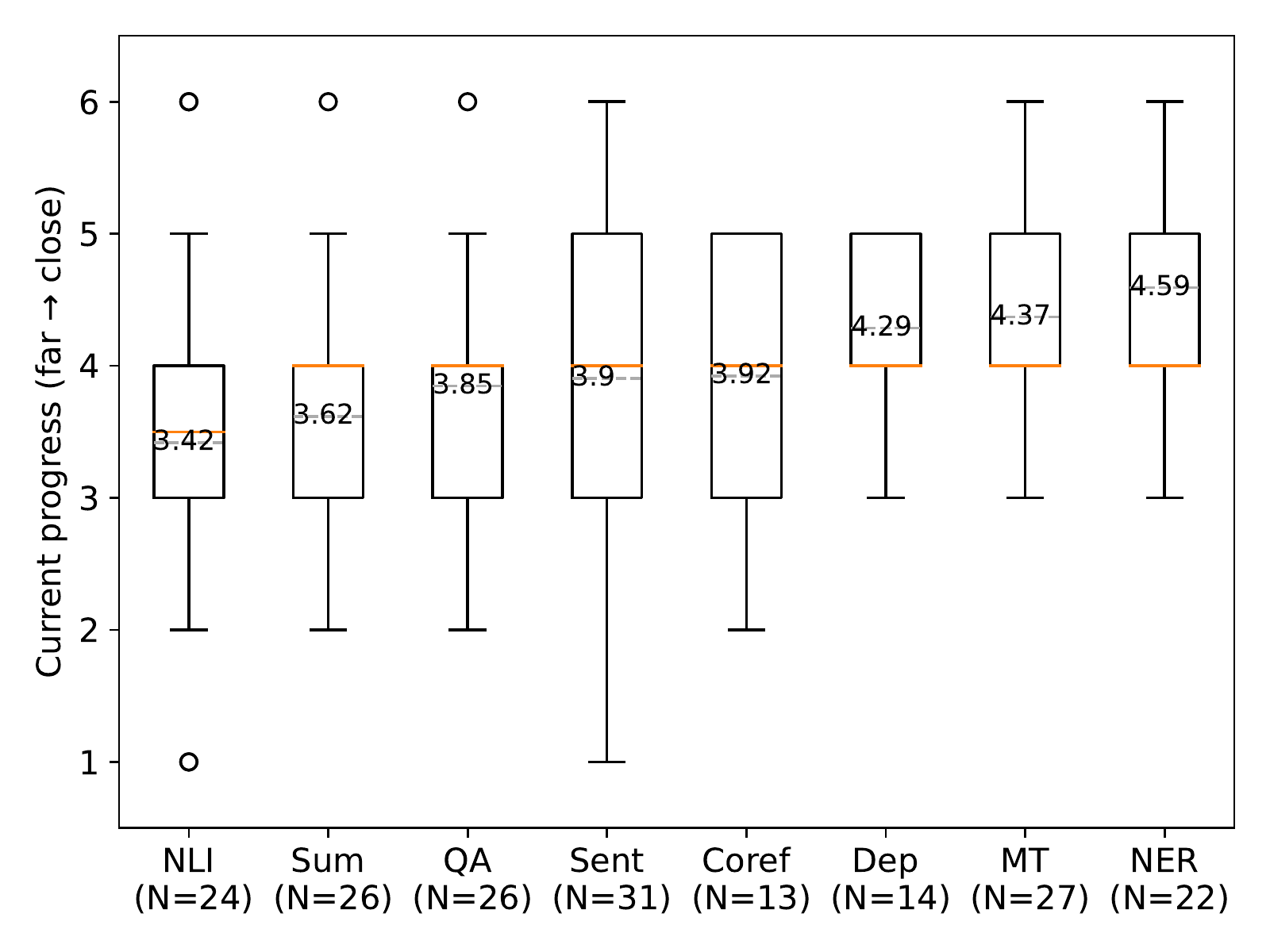}
        \caption{Perceived current progress among survey participants who consider themselves to have ``passing knowledge'' about each task.}
        \label{fig:current_progress_novice}
    \end{subfigure}
    \caption{}
    \label{fig:current_progress}
\end{figure}
\FloatBarrier

\newpage

\subsection{Potential progress}
\begin{figure}[!ht]
    \centering
    \begin{subfigure}[]{0.5\textwidth}
        \includegraphics[width=\textwidth]{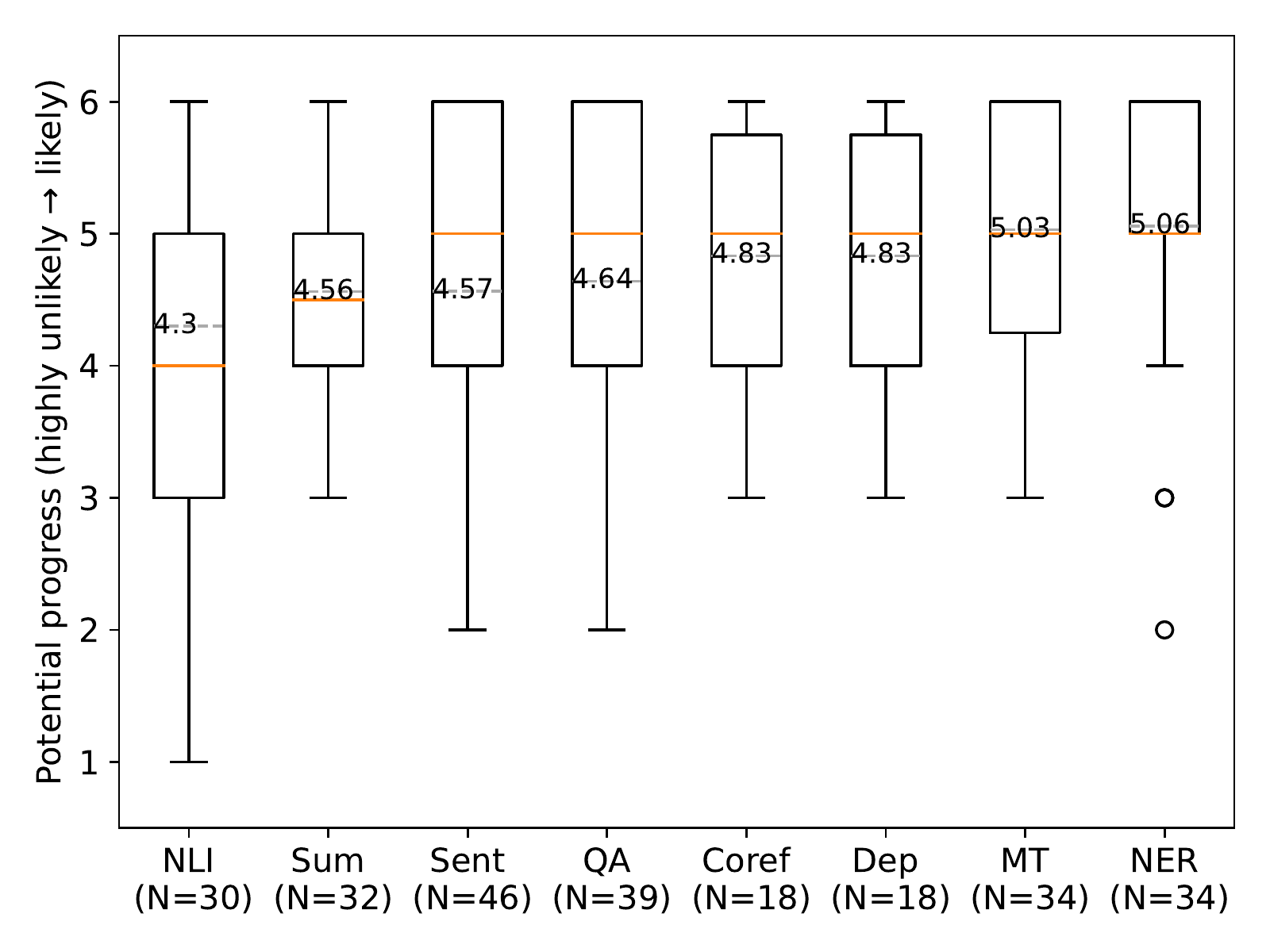}
        \caption{Perceived potential progress among all responding practitioners for each task.}
        \label{fig:potential_progress_all}
    \end{subfigure}
    \begin{subfigure}[]{0.5\textwidth}
        \includegraphics[width=\textwidth]{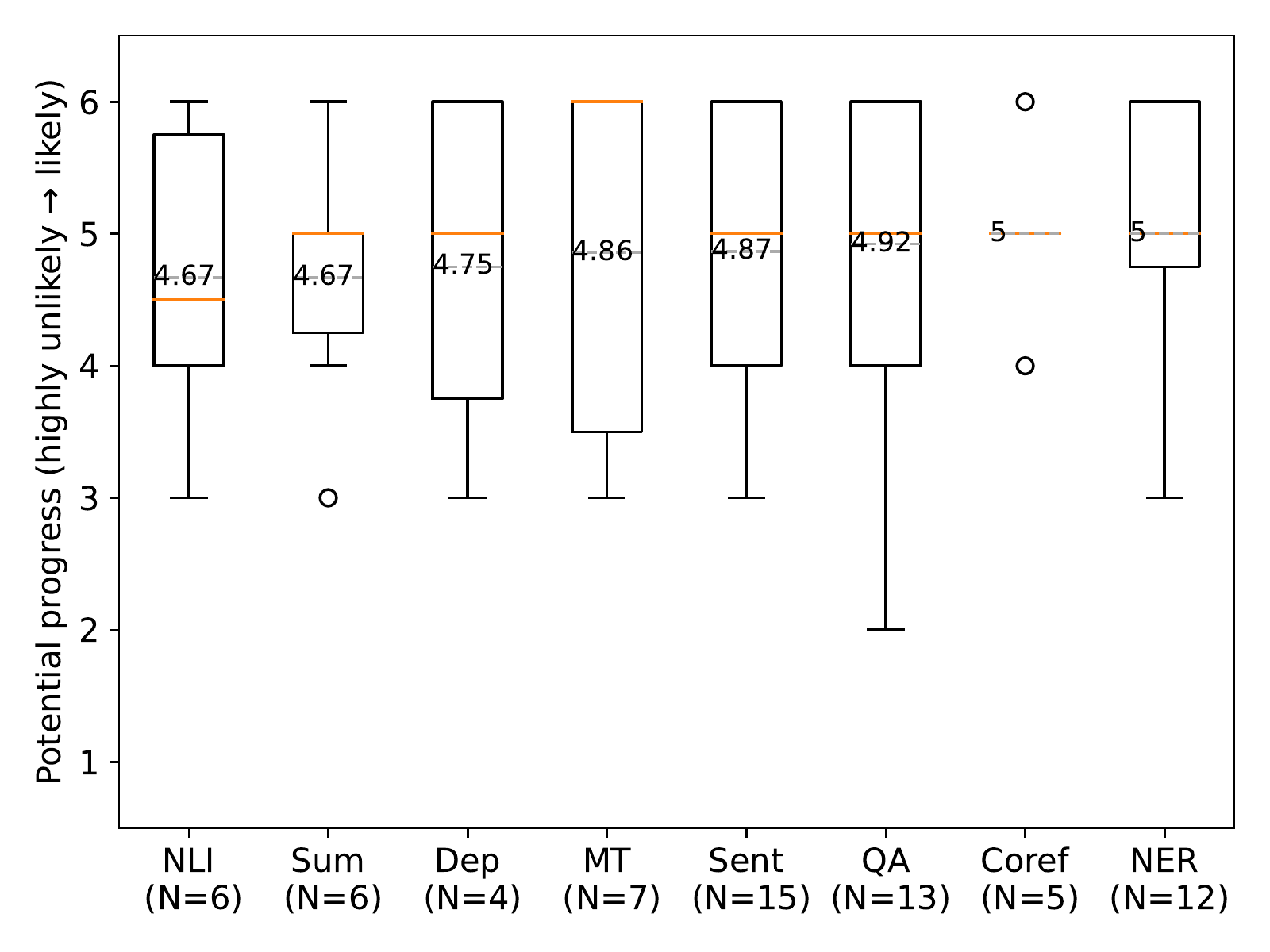}
        \caption{Perceived potential progress among survey participants who consider themselves an ``expert'' at each task.}
        \label{fig:potential_progress_expert}
    \end{subfigure}
    \begin{subfigure}[b]{0.5\textwidth}
        \includegraphics[width=\textwidth]{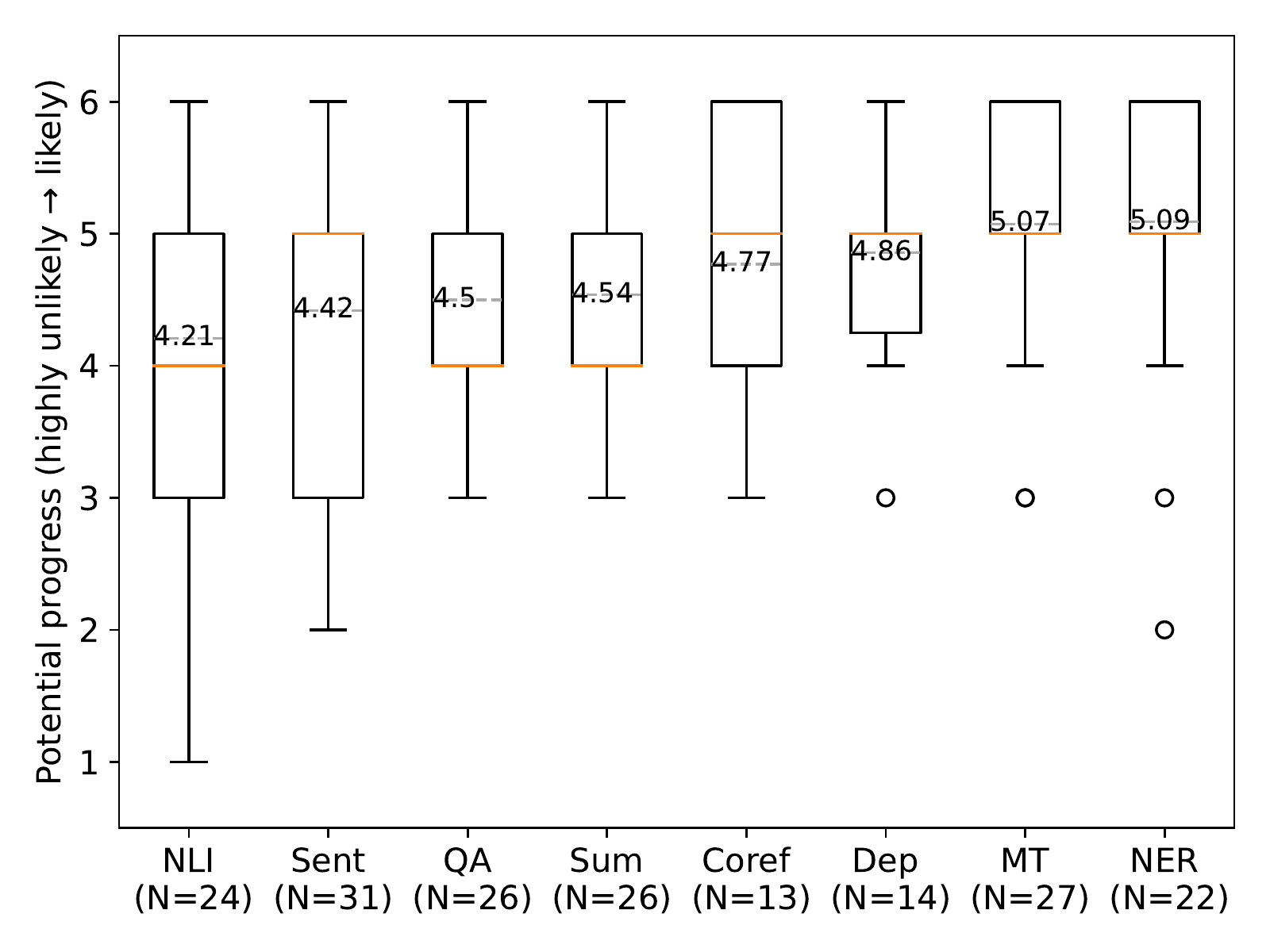}
        \caption{Perceived potential progress among survey participants who consider themselves to have ``passing knowledge'' about each task.}
        \label{fig:potential_progress_novice}
    \end{subfigure}
    \caption{}
    \label{fig:potential_progress}
\end{figure}
\FloatBarrier

\newpage

\section{Model Capabilities Purportedly Assessed by Benchmarks}
\label{sec:model_capabilities}

\subsection{Benchmark Review Protocol}

Our use of all the benchmarks below was reviewed by an IRB. The IRB deemed that our analysis of the benchmarks was not prohibited by their license nor the terms of use of the benchmark data sources. The IRB also confirmed that the benchmarks do not contain any information that names or uniquely identifies individual people or offensive content.

\subsection{Benchmark Analysis}

\begin{table}[!ht]
    \centering
    \begin{tabular}{|p{0.25\textwidth}|p{0.7\textwidth}|}
        \toprule
        \textbf{Capability} & \textbf{Benchmarks} \\
        \midrule
        Capture meaning & \textbf{SST \cite{socher-etal-2013-recursive}}: ``capture the meaning of longer phrases''
        \newline
        \newline
        \textbf{IMDb \cite{Maas2011LearningWV}:}  ``capture both semantic and sentiment similarities among words,'' ``learns [\ldots] nuanced sentiment information'' 
        \\
        \midrule
        Outperform humans & \textbf{Cornell movie reviews \cite{Pang2002ThumbsUS}:} ``outperform human-produced baselines'' \\
        \midrule
        Handle linguistic phenomena & \textbf{SST \cite{socher-etal-2013-recursive}}: ``presents new challenges for sentiment compositionality,'' ``capture the effects of negation,'' ``capture complex linguistic phenomena,'' ``learn that sentiment of phrases following the contrastive conjunction ‘but’ dominates,'' ``from a linguistic or cognitive standpoint, ignoring word order in the treatment of a semantic task is not plausible''
        \newline
        \newline
        \textbf{Cornell movie reviews \cite{Pang2002ThumbsUS}:} handle ``thwarted-expectations rhetorical device'' \\
        \midrule 
        Handle phenomena in real-world data & \textbf{SST \cite{socher-etal-2013-recursive}}: ``there is a need to better capture sentiment from short comments, such as Twitter data'' \\
         \bottomrule
    \end{tabular}
    \caption{Model capabilities that \textsc{Sent} benchmarks are intended to measure.}
    \label{tab:capabilities_measured_sent}
\end{table}
\FloatBarrier

\begin{table}[!ht]
    \centering
    \begin{tabular}{|p{0.25\textwidth}|p{0.7\textwidth}|}
        \toprule
        \textbf{Capability} & \textbf{Benchmarks} \\
        \midrule
        Understand language & \textbf{MNLI \cite{williams-etal-2018-broad}:} ``evaluation of methods for sentence understanding''
        \newline
        \newline
        \textbf{SNLI \cite{Bowman2015ALA}:} ``understanding entailment and contradiction is fundamental to understanding natural language,'' ``models' attempts to shortcut this kind of inference through lexical cues can lead them astray''
        \newline
        \newline
        \textbf{XNLI \cite{conneau-etal-2018-xnli}:} ``[test bed for] crosslingual language understanding''
        \\
        \midrule
        Handle linguistic phenomena & \textbf{MNLI \cite{williams-etal-2018-broad}:} ``handle phenomena like lexical entailment, quantification, coreference, tense, belief, modality, and lexical and syntactic ambiguity''
        \newline
        \newline
        \textbf{SNLI \cite{Bowman2015ALA}:} ``NLI is an ideal testing ground for theories of semantic representation,'' ``additional attention to compositional semantics would pay off''
        \newline
        \newline
        \textbf{RTE \cite{Giampiccolo2008TheFP}:} ``finding equivalences and similarities at lexical, syntactic and semantic levels'' \\
        \midrule
        Handle various domains & \textbf{MNLI \cite{williams-etal-2018-broad}:} ``corpus [\ldots] meant to approximate full diversity of ways in which modern standard American English is used,'' ``represents both written and spoken speech in a wide range of styles, degrees of formality, and topics,'' ``benchmark for cross-genre domain adaptation''
        \newline
        \newline
        \textbf{SNLI \cite{Bowman2015ALA}:} ``evaluation of domain-general approaches'' \\
        \midrule
        Possess benchmark-external knowledge & \textbf{MNLI \cite{williams-etal-2018-broad}:}  ``using only [\ldots] what you know about the world''
        \newline
        \newline
        \textbf{SNLI \cite{Bowman2015ALA}:} ``data collected draws fairly extensively on commonsense knowledge'' \\
        \midrule
        Aid in other NLP tasks &
        \textbf{XNLI \cite{conneau-etal-2018-xnli}:} ``evaluation of pretrained general-purpose language universal sentence encoders''
        \newline
        \newline
        \textbf{RTE \cite{Giampiccolo2008TheFP}:} ``captures major semantic inference needs across many natural language processing applications, such as Question Answering (QA), Information Retrieval (IR), Information Extraction (IE), and multi-document summarization (SUM)'' \\
        \bottomrule
    \end{tabular}
    \caption{Model capabilities that \textsc{NLI} benchmarks are intended to measure.}
    \label{tab:capabilities_measured_nli}
\end{table}
\FloatBarrier

\begin{table}[!ht]
    \centering
    \resizebox{0.8\textwidth}{!}{\begin{tabular}{|p{0.25\textwidth}|p{0.7\textwidth}|}
        \toprule
        \textbf{Capability} & \textbf{Benchmarks} \\
        \midrule
        Understand language & \textbf{SQuAD \cite{Rajpurkar2016SQuAD1Q}:} ``requiring both understanding of natural language and knowledge about the world,'' ``towards the end goal of natural language understanding''
        \newline
        \newline
        \textbf{HotpotQA \cite{yang-etal-2018-hotpotqa}:}
        ``test their understanding of both language and common concepts such as numerical magnitude''
        \\
        \midrule
        Handle linguistic phenomena & \textbf{SQuAD \cite{Rajpurkar2016SQuAD1Q}:} ``all examples have some sort of lexical or syntactic divergence between the question and the answer in the passage''
        \newline
        \newline 
        \textbf{TriviaQA \cite{joshi-etal-2017-triviaqa}:} ``has relatively complex, compositional questions,'' ``has considerable syntactic and lexical variability'' \\
        \midrule
        Reason over a context & \textbf{SQuAD \cite{Rajpurkar2016SQuAD1Q}:} ``multiple sentence reasoning''
        \newline
        \newline
        \textbf{HotpotQA \cite{yang-etal-2018-hotpotqa}:} ``questions require finding and reasoning over multiple supporting documents to answer,'' ``test the reasoning ability of intelligent systems,'' numerous ``types of multi-hop reasoning required to answer questions''
        \newline
        \newline
        \textbf{TriviaQA \cite{joshi-etal-2017-triviaqa}:} ``requires more cross sentence reasoning to find answers'' \\
        \midrule
        Possess benchmark-external knowledge & \textbf{SQuAD \cite{Rajpurkar2016SQuAD1Q}:} ``requires both understanding of natural language and knowledge about the world''
        \newline
        \newline
        \textbf{HotpotQA \cite{yang-etal-2018-hotpotqa}:} ``the questions are [\ldots] not constrained to any pre-existing knowledge bases or knowledge schemas''
        \newline
        \newline
        \textbf{TriviaQA \cite{joshi-etal-2017-triviaqa}:} ``17\% of the examples required some form of world knowledge'' \\
        \midrule 
        Handle various domains & \textbf{HotpotQA \cite{yang-etal-2018-hotpotqa}:} ``the questions are [\ldots] not constrained to any pre-existing knowledge bases or knowledge schemas,'' ``our dataset covers a diverse variety of questions centered around entities, locations, events, dates, and numbers, as well as yes/no questions directed at comparing two entities'' 
        \newline
        \newline
        \textbf{TriviaQA \cite{yang-etal-2018-hotpotqa}:} models ``should be able to deal with large amount of text from various sources such as news articles, encyclopedic entries and blog articles'' \\
        \midrule
        Handle phenomena in real-world data & \textbf{SQuAD \cite{Rajpurkar2016SQuAD1Q}:} ``existing datasets for RC [\ldots] that are large [\ldots] are semi-synthetic and do not share the same characteristics as explicit reading comprehension questions''
        \newline
        \newline
        \textbf{TriviaQA \cite{joshi-etal-2017-triviaqa}:} ``first dataset where full-sentence questions are authored organically'' \\ %
        \midrule 
        be on par with humans & \textbf{SQuAD \cite{Rajpurkar2016SQuAD1Q}:} ``these results are still well behind human performance''
        \newline
        \newline
        \textbf{HotpotQA \cite{yang-etal-2018-hotpotqa}:} ``if the baseline model were provided with the correct supporting paragraphs to begin with, it achieves parity with the crowd worker in finding supporting facts'' 
        \newline
        \newline
        \textbf{TriviaQA \cite{joshi-etal-2017-triviaqa}:} ``neither approach comes close to human performance'' \\
        \bottomrule
    \end{tabular}}
    \caption{Model capabilities that \textsc{QA} benchmarks are intended to measure.}
    \label{tab:capabilities_measured_qa}
\end{table}
\FloatBarrier

\begin{table}[!ht]
    \centering
    \begin{tabular}{|p{0.25\textwidth}|p{0.7\textwidth}|}
        \toprule
        \textbf{Capability} & \textbf{Benchmarks} \\
        \midrule
        Think & \textbf{Winograd Schema Challenge \cite{Levesque2011TheWS}:} ``thinking is required to get a correct answer with high probability'' \\
        \midrule
        Handle linguistic phenomena & \textbf{Winograd Schema Challenge \cite{Levesque2011TheWS}:}
        ``question involves determining the referent of the pronoun or possessive adjective'' \\
        \midrule
        be on par with humans & \textbf{Winograd Schema Challenge \cite{Levesque2011TheWS}:} ``required to achieve human-level accuracy in choosing the correct disambiguation'' \\
        \midrule
        Possess benchark-external knowledge & \textbf{Winograd Schema Challenge \cite{Levesque2011TheWS}:} ``you need to have background knowledge that is not expressed in the words of the sentence'' \\
        \midrule
        Aid in other NLP tasks & \textbf{Winograd Schema Challenge \cite{Levesque2011TheWS}:} ``it is sometimes possible to find sentences in natural text that can easily be turned into Winograd schemas'' \\
        \midrule
        Handle various genres & \textbf{OntoNotes \cite{Hovy2006OntoNotesT9}:} ``annotation will cover [\ldots] multiple genres (newswire, broadcast news, news groups, weblogs, etc.), to create a resource that is broadly applicable'' \\
        \bottomrule
    \end{tabular}
    \caption{Model capabilities that \textsc{Coref} benchmarks are intended to measure.}
    \label{tab:capabilities_measured_coref}
\end{table}
\FloatBarrier

\begin{table}[!ht]
    \centering
    \begin{tabular}{|p{0.25\textwidth}|p{0.7\textwidth}|}
        \toprule
        \textbf{Capability} & \textbf{Benchmarks} \\
        \midrule
        Understand language & \textbf{XSum \cite{Narayan2018DontGM}:} ``posing several challenges relating to understanding (i.e., identifying important content),'' ``high-level document knowledge in terms of topics and long-range dependencies is critical for recognizing pertinent content and generating informative summaries''
        \newline
        \newline
        \textbf{CNN/DM \cite{Nallapati2016AbstractiveTS}:} ``capturing the `meaning' of complex sentences'' \\
        \midrule
        Generate novel language & \textbf{XSum \cite{Narayan2018DontGM}:} ``generation (i.e., aggregating and rewording the identified content into a summary),'' ``there are [\ldots] novel [$n$-]grams in the XSum reference summaries'' \\
        \midrule
        Handle linguistic phenomena & \textbf{XSum \cite{Narayan2018DontGM}:} ``displays multiple levels of abstraction including paraphrasing, fusion, synthesis, and inference'' \\
        \midrule
        Handle various domains & \textbf{XSum \cite{Narayan2018DontGM}:} ``collected 226,711 Wayback archived BBC articles ranging over almost a decade (2010 to 2017) and covering a wide variety of domains'' \\
        \midrule
        Possess benchmark-external knowledge & \textbf{CNN/DM \cite{Nallapati2016AbstractiveTS}:} ``potentially using vocabulary unseen in the source document'' \\
        \bottomrule
    \end{tabular}
    \caption{Model capabilities that \textsc{Sum} benchmarks are intended to measure.}
    \label{tab:capabilities_measured_sum}
\end{table}
\FloatBarrier

\begin{table}[!ht]
    \centering
    \begin{tabular}{|p{0.25\textwidth}|p{0.7\textwidth}|}
        \toprule
        \textbf{Capability} & \textbf{Benchmarks} \\
        \midrule
        Handle various domains & \textbf{OntoNotes \cite{Hovy2006OntoNotesT9}:} ``annotation will cover [\ldots] multiple genres (newswire, broadcast news, news groups, weblogs, etc.), to create a resource that is broadly applicable'' \\
        \midrule
        Handle various languages & \textbf{CoNLL-2003 \cite{Sang2003IntroductionTT}:} ``language-independent named entity recognition'' \\
        \midrule
        Aid in real-world applications of task & \textbf{CoNLL-2003 \cite{Sang2003IntroductionTT}:} ``named entity recognition is an important task of information extraction systems'' \\
        \midrule 
        Possess benchmark-external knowledge & \textbf{CoNLL-2003 \cite{Sang2003IntroductionTT}:} `` interested in approaches that made use of resources other than the supplied training data'' \\
        \bottomrule
    \end{tabular}
    \caption{Model capabilities that \textsc{NER} benchmarks are intended to measure.}
    \label{tab:capabilities_measured_ner}
\end{table}
\FloatBarrier

\begin{table}[!ht]
    \centering
    \begin{tabular}{|p{0.25\textwidth}|p{0.7\textwidth}|}
        \toprule
        \textbf{Capability} & \textbf{Benchmarks} \\
        \midrule
        Understand language & \textbf{Penn Treebank \cite{Marcus1993BuildingAL}:} ``progress can be made in both text understanding and spoken language understanding'' \\
        \midrule
        Handle phenomena in real-world data & \textbf{Penn Treebank \cite{Marcus1993BuildingAL}:} ``in naturally occurring unconstrained materials'' \\
        \midrule
        Handle linguistic phenomena & \textbf{Penn Treebank \cite{Marcus1993BuildingAL}:} ``evaluation and comparison of the adequacy of parsing models'' \\
        \midrule
        Handle various languages & \textbf{Universal Dependencies \cite{nivre-etal-2016-universal}:} ``facilitate multilingual natural language processing'' \\
        \midrule 
        Handle various genres & \textbf{Universal Dependencies \cite{nivre-etal-2016-universal}:} ``most treebanks are constituted of different genres'' \\
        \bottomrule
    \end{tabular}
    \caption{Model capabilities that \textsc{Dep} benchmarks are intended to measure.}
    \label{tab:capabilities_measured_dep}
\end{table}
\FloatBarrier

\begin{table}[!ht]
    \centering
    \begin{tabular}{|p{0.25\textwidth}|p{0.7\textwidth}|}
        \toprule
        \textbf{Capability} & \textbf{Benchmarks} \\
        \midrule
        Handle various languages & \textbf{Europarl \cite{Koehn2005EuroparlAP}:} ``parallel text in 11 languages''
        \newline
        \newline
        \textbf{WMT-2007 \cite{callison-burch-etal-2007-meta}:} ``translating French, German, Spanish, and Czech to English and back''\\
        \midrule
        Handle linguistic phenomena & \textbf{Europarl \cite{Koehn2005EuroparlAP}:} ``reason for the difficulty of translating into a language is morphological richness'' \\
        \midrule
        Generate fluent and adequate language & \textbf{WMT-2007 \cite{callison-burch-etal-2007-meta}:} ``fluency and adequacy''
        \newline
        \newline
        \textbf{OpenMT\tablefootnote{\url{https://www.nist.gov/itl/iad/mig/open-machine-translation-evaluation}}:} ``goal is for the output to be an adequate and fluent translation of the original'' \\
        \bottomrule
    \end{tabular}
    \caption{Model capabilities that \textsc{MT} benchmarks are intended to measure.}
    \label{tab:capabilities_measured_wmt}
\end{table}
\FloatBarrier

\newpage

\section{Additional Examples of Essentially Contested Constructs}
\label{sec:essentially_contested_examples}

\begin{itemize}[noitemsep,topsep=0pt,parsep=0pt,partopsep=0pt,leftmargin=*]
    \item \textbf{Understanding language:} Practitioners often do not explain how they conceptualize language understanding, nor do they address disagreement about whether models are capable of understanding language or language can be understood from text alone \cite{metasurvey2022}. %
    \item \textbf{Handling real-world phenomena:} Practitioners often leave the definition of ``real-world'' open-ended, despite foregrounding certain domains in their conceptualization of ``real-world,'' and do not address contention around the feasibility of capturing ``everything in the whole wide world'' \cite{Raji2021AIAT}. 
\end{itemize}

\section{Additional Disagreements in NLP Task Conceptualization}
\label{sec:additional_disagreements}

\newcolumntype{P}[1]{>{\centering\arraybackslash}p{#1}}
\begin{table}[!ht]
    \centering
    \begin{adjustbox}{max width=\textwidth}
    \begin{tabular}{|p{0.07\textwidth}|p{0.5\textwidth}|p{0.43\textwidth}|}
         \toprule
         \thead{\textbf{Task}} & \thead{\textbf{Disagreement in conceptualization?}} & \thead{\textbf{Conceptualization disagreement examples}} \\
         \midrule
        \multirow{6}{=}{\textsc{Sent}} & $C_\tau$: yes (\Cref{tab:capabilities_measured_sent})
                        & \multirow{6}{=}{SST, IMDb, Cornell movie reviews datasets operationalize sentiment with single gold label \cite{socher-etal-2013-recursive, Maas2011LearningWV, Pang2002ThumbsUS}}
                        \\
                        \cline{2-2}
                       & $y_\tau$: yes; lack of ``real 'ground truth''' due to differing perceptions of sentiment \cite{ovesdotter-alm-2011-subjective, Hagerer2021EndtoEndAB} & \\
                       \cline{2-2}
                       & $E_\tau$: yes; \{sentiment, capture meaning, outperform humans\} $\subset E_\tau$ (\Cref{tab:capabilities_measured_sent}) & \\
         \midrule   
         \multirow{5}{=}{\textsc{NER}} & $C_\tau$: yes (\Cref{tab:capabilities_measured_ner})
                            & \multirow{5}{=}{OntoNotes, CoNLL-2003 datasets operationalize type of ambiguous entities with single gold label \cite{Hovy2006OntoNotesT9, Sang2003IntroductionTT}}
                            \\
                        \cline{2-2}
                        & $y_\tau$: yes; inherent semantic ambiguity induces disagreement about $y_\tau$ \cite{Zhang2013NamedER} & \\
                        \cline{2-2}
                        & $E_\tau$: yes; \{ possess benchmark-external knowledge \} $\subset E_\tau$ (\Cref{tab:capabilities_measured_ner}) & \\
         \midrule
         \multirow{6}{=}{\textsc{Dep}} & $C_\tau$: yes (\Cref{tab:capabilities_measured_dep})
                        & \multirow{6}{=}{benchmarks make use of inconsistent annotation formats due to differing conceptualizations of parsing \cite{Dredze2007FrustratinglyHD, nivre-etal-2016-universal}}
                        \\
                        \cline{2-2}
                        & $y_\tau$: yes; syntactic ambiguity \cite{Ackerman2015InfluencesOP, keith-etal-2018-monte} and systematic disagreement about parts of speech \cite{plank-etal-2014-linguistically} yield differing $y_\tau$ & \\
                        \cline{2-2}
                        & $E_\tau$: yes; \{ understand language, handle phenomena in real-world data \} $\subset E_\tau$ (\Cref{tab:capabilities_measured_dep}) & \\
         \midrule
         \multirow{8}{=}{\textsc{MT}} & $C_\tau$: yes (\Cref{tab:capabilities_measured_wmt})
                        & \multirow{8}{=}{Europarl, WMT-2007 datasets contain single reference translations \cite{Koehn2005EuroparlAP, callison-burch-etal-2007-meta}}
                        \\
                        \cline{2-2}
                        & $y_\tau$: adequacy of translations in $y_\tau$ is subjective \cite{white-oconnell-1993-evaluation}; can be unclear how to translate lexical and syntactic ambiguity in source language \cite{Pericliev1984HandlingSA, Baker1994CopingWA}, or translate from language without to with grammatical gender \cite{gonen-webster-2020-automatically} & \\
                        \cline{2-2}
                         & $E_\tau$: yes; \{ fluency, adequacy \} $\subset E_\tau$ (\Cref{tab:capabilities_measured_wmt}) & \\
        \bottomrule
    \end{tabular}
    \end{adjustbox}
    \caption{Additional disagreements in the conceptualization of NLP tasks and examples of resultant conceptualization disagreements.}
    \label{tab:additional_findings_summary}
\end{table}

\begin{figure}[!ht]
\centering
\resizebox{0.5\columnwidth}{!}{\begin{tabular}{ll}
\toprule
\textbf{Example} & \textit{Sentences} \\
\textit{Premise} & Isn't a woman's body \\
& her most personal property? \\
\textit{Hypothesis} & Isn't a woman's body \\
& sacred property? \\
\textit{Annotator labels} & E, E, E, N, N \\
\textit{Gold label} & Entailment \\
\midrule
\textbf{Issues} & \textit{Description} \\
\textit{Conceptualization} &
    \tabitem unclear whether a question can \\
    & \hphantom{\tabitem}entail or contradict any hypothesis\footnotemark \\
    & \tabitem unclear whether any premise can \\
    & \hphantom{\tabitem}entail or contradict a question \\
\bottomrule
\end{tabular}}
\caption{Example test instance from the MNLI benchmark  \cite{williams-etal-2018-broad}, accompanied by issues with the conceptualization of MNLI that the instance reflects. The \textsc{NLI} task captures whether the premise entails (E), contradicts (C), or is neutral (N) with respect to the hypothesis.}
\label{fig:mnli_example}
\end{figure}
\FloatBarrier
\footnotetext{Although not stated in the paper, according to \citet{williams-etal-2018-broad}, a question entails the set of its possible answers.}

\end{document}